\documentclass[letterpaper]{article} % DO NOT CHANGE THIS
\usepackage[]{aaai25}
\usepackage{times}  % DO NOT CHANGE THIS
\usepackage{helvet}  % DO NOT CHANGE THIS
\usepackage{courier}  % DO NOT CHANGE THIS
\usepackage[hyphens]{url}  % DO NOT CHANGE THIS
\usepackage{graphicx} % DO NOT CHANGE THIS
\usepackage{natbib}  % DO NOT CHANGE THIS AND DO NOT ADD ANY OPTIONS TO IT
\usepackage{caption} % DO NOT CHANGE THIS AND DO NOT ADD ANY OPTIONS TO IT
\usepackage{algorithm}
\usepackage{algorithmic}

\usepackage[T1]{fontenc}
\usepackage[utf8]{inputenc}
\usepackage{microtype}

\usepackage{inconsolata}
\usepackage{filecontents}

\newcommand{\ap}[1]{\textbf{\textcolor{blue}{#1}}}
\usepackage{amsmath}
\usepackage{lipsum}

\usepackage{epigraph}
\usepackage{svg}

\usepackage{booktabs}
\usepackage{tikz}
\usepackage{pgfplots}
\usepgfplotslibrary{external}
\usepackage{multirow}
\usepackage{enumitem}

\usepgfplotslibrary{colorbrewer}
\pgfplotsset{compat = 1.15, cycle list/Set1-8} 
\usetikzlibrary{pgfplots.statistics, pgfplots.colorbrewer} 
\usepackage{pgfplotstable}
\usepackage{filecontents}
\usepackage{varwidth}
\usepackage{soul}
\usepackage{epigraph}

\usetikzlibrary{pgfplots.groupplots}
\usetikzlibrary{shapes.geometric}
\usetikzlibrary{positioning,fit,shapes.geometric,backgrounds, calc}
\usetikzlibrary{arrows,decorations.markings}
\usetikzlibrary{shapes.arrows}
\usetikzlibrary{patterns}
\tikzset{textnode/.style={inner sep=0pt,outer sep=0,execute at begin node={\strut}}}
\tikzstyle{state} = [textnode,circle, draw, inner sep=0pt, outer sep=0]
\usepgfplotslibrary{groupplots}
\pgfplotsset{every axis/.append style={
                    xlabel={$x$},          % default put x on x-axis
                    ylabel={$y$},          % default put y on y-axis
                    label style={font=\sffamily},
                    tick label style={font=\sffamily\footnotesize},
                    xticklabel style = {font=\sffamily\footnotesize},
                    title style = {font=\normalsize\sffamily},
                    ylabel near ticks,
                    y label style={font=\sffamily\footnotesize},
                    xlabel near ticks,
                    x label style={font=\sffamily\footnotesize},
                    legend cell align={left},
                    legend style={draw=none, font=\sffamily\scriptsize},
                    },
                    legend image code/.code={
                    \draw[mark repeat=2,mark phase=2]
                        plot coordinates {
                        (0cm,0cm)
                        (0.15cm,0cm)        %% default is (0.3cm,0cm)
                        (0.3cm,0cm)         %% default is (0.6cm,0cm)
                        };%
                    }
                    }
\pgfplotsset{compat=newest}

\makeatletter
\pgfplotsset{
    boxplot prepared from table/.code={
        \def\tikz@plot@handler{\pgfplotsplothandlerboxplotprepared}%
        \pgfplotsset{
            /pgfplots/boxplot prepared from table/.cd,
            #1,
        }
    },
    /pgfplots/boxplot prepared from table/.cd,
        table/.code={\pgfplotstablecopy{#1}\to\boxplot@datatable},
        row/.initial=0,
        make style readable from table/.style={
            #1/.code={
                \pgfplotstablegetelem{\pgfkeysvalueof{/pgfplots/boxplot prepared from table/row}}{##1}\of\boxplot@datatable
                \pgfplotsset{boxplot/#1/.expand once={\pgfplotsretval}}
            }
        },
        make style readable from table=lower whisker,
        make style readable from table=upper whisker,
        make style readable from table=lower quartile,
        make style readable from table=upper quartile,
        make style readable from table=median,
        make style readable from table=lower notch,
        make style readable from table=upper notch,
                make style readable from table=average
}
\makeatother

\newcommand{\computerowindex}[2]{
\def\xindex{-1}
\pgfplotstableforeachcolumnelement{x}\of#1\as\cell{%
\edef\cellname{#2}
    \ifx\cell\cellname\let\xindex\pgfplotstablerow\fi   
}
}
\newcommand{\boxplotprepared}[2]{
\computerowindex{#1}{#2}
\expandafter\boxplotpreparedNum\expandafter#1\expandafter{\xindex}
}

\newcommand{\boxplotpreparedNum}[2]{
\addplot[boxplot prepared from table={
    table=#1,
    row=#2,
    average=avg,
    lower whisker = firstmin,
    upper whisker = firstmax,
    lower quartile = q1,
    upper quartile = q3,
    median=q2,
    },
    boxplot prepared]
coordinates {};
}

\newcommand*{\eg}{{\em e.g.}}

\newcommand*{\ie}{{\em i.e.}}
\newcommand{\rev}[1]{\textcolor{black}{#1}} 
\newcommand{\answerYes}[1]{\textcolor{blue}{#1}} 
 
\newcommand{\answerNA}[1]{\textcolor{gray}{#1}}

\definecolor{conflict}{RGB}{215,48,39}        % Crimson Red
\definecolor{econ}{RGB}{253,174,97}          % Warm Gold/Orange
\definecolor{episodic}{RGB}{44,123,182}      % Indigo Blue
\definecolor{strategy}{RGB}{49,163,84}       % Teal Green
\definecolor{human}{RGB}{117,107,177}        % Soft Purple
\definecolor{thematic}{RGB}{102,102,102}     % Slate Gray

\definecolor{extremeleft}{RGB}{44,84,148}
\definecolor{left}{RGB}{80,130,200}
\definecolor{leftcenter}{RGB}{145,165,220}
\definecolor{center}{RGB}{165,135,190}
\definecolor{rightcenter}{RGB}{220,150,150}
\definecolor{right}{RGB}{200,100,100}
\definecolor{extremeright}{RGB}{150,50,50}

\definecolor{low}{RGB}{198,80,80}
\definecolor{mixed}{RGB}{220,160,60}
\definecolor{high}{RGB}{100,170,100}
\definecolor{veryhigh}{RGB}{70,130,180}
\usepackage{newfloat}
\usepackage{listings}
\DeclareCaptionStyle{ruled}{labelfont=normalfont,labelsep=colon,strut=off} % DO NOT CHANGE THIS
\floatstyle{ruled}
\newfloat{listing}{tb}{lst}{}
\floatname{listing}{Listing}
\title{Social and Political Framing in Search Engine Results}
\author{
    Amrit Poudel, Tim Weninger
}
\affiliations{
    University of Notre Dame
    \{apoudel, tweninger\}@nd.edu
}

\begin{document}

\maketitle

\begin{abstract}
Search engines play a crucial role in shaping public discourse by influencing how information is accessed and framed. While prior research has extensively examined various dimensions of search bias---such as content prioritization, indexical bias, political polarization, and sources of bias---an important question remains underexplored: how do search engines and ideologically-motivated user queries contribute to bias in search results. This study analyzes the outputs of major search engines using a dataset of political and social topics. The findings reveal that search engines not only prioritize content in ways that reflect underlying biases but also that ideologically-driven user queries exacerbate these biases, resulting in the amplification of specific narratives. Moreover, significant differences were observed across search engines in terms of the sources they prioritize. These results suggest that search engines may play a pivotal role in shaping public perceptions by reinforcing ideological divides, thereby contributing to the broader issue of information polarization.
\end{abstract}

\section{Introduction}

\epigraph{
``There is competition for those top ten seats. There is serious competition. People are trying to take away the top spots every day. They are always trying to fine-tune and tweak their HTML code and learn the next little trick. The best players even know dirty ways to ``bump off'' their competition while protecting their own sites.''
}{
~\cite{anderson1997hits}
    % -- Anderson \& Henderson, 1997
}

Every day, billions of people turn to search engines like Google, Bing and DuckDuckGo for information, making these platforms indispensable in the digital age. With billions of queries submitted to search engines daily, they have become the modern-day equivalent of libraries---vast directories of human knowledge curated and presented at the click of a button. In many ways, they also serve as media companies~\cite{goldman2005search}. Google, one of the most widely used search engines, claims to rank results based on the following criteria\footnote{\url{https://www.google.com/intl/en/search/howsearchworks/how-search-works/ranking-results/}}: relevance, quality, usability, and context. These criteria align closely with the four components of search engine quality assessment \ie, index quality, results quality, search features quality, and search engine usability~\cite{lewandowski2008web}. However, the impartiality of these algorithms has been increasingly called into question by political leaders, organizations, and scholars, raising concerns about the suppression and promotion of content~\cite{kulshrestha2019search, gezici2021evaluation}.

\rev{Nevertheless, users often trust search engine results~\cite{pan2007google} and treat top-ranked links as more credible. This gap between general awareness and everyday behavior highlights the need to examine how search systems reinforce or counter ideological framing.}

\begin{figure*}
    \centering
    \includegraphics[width=\linewidth]{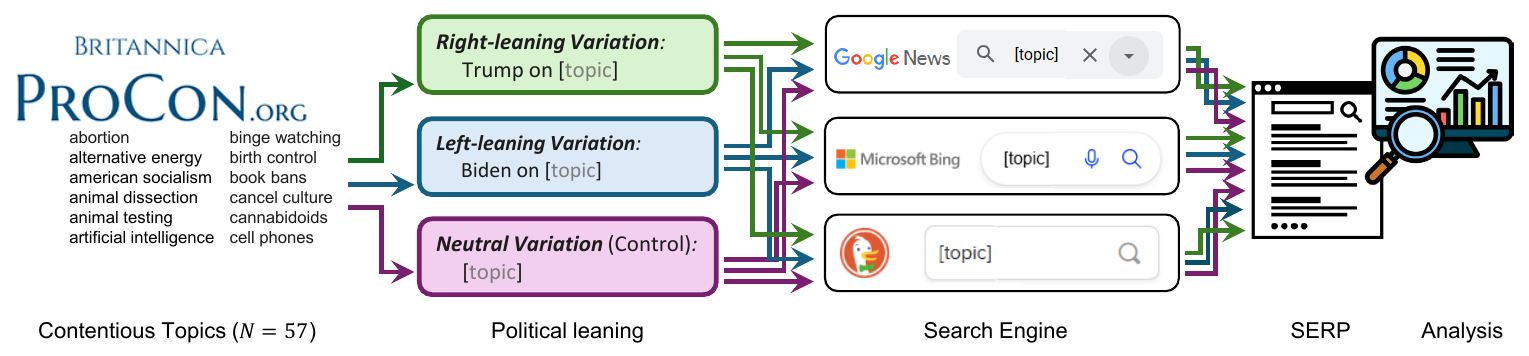}
    \caption{Flowchart illustrating the analysis of 57 debate topics from ProCon.org. Each query was tested in three variations: Right, Left, and Neutral and submitted to the news sections of three search engines. The retrieved results were aggregated and analyzed for potential variations in semantics, frames, source dynamics, ideological bias and factuality.}
    \label{fig:method}
\end{figure*}

\rev{Before the rise of search engines, traditional media such as newspapers, radio, and television served as central sources of information. Though once celebrated as pillars of democracy, these institutions have long been shaped by commercial and political interests~\cite{introna2000shaping}. Editorial decisions (\eg, about what to print and how to print it) significantly influence how news is selected and framed, shaping public perception of political issues~\cite{bush1951analysis}. Search engines, while initially praised for democratizing access to information, are subject to similar biases~\cite{poudel2024navigating, poudel2025digital}. Ranking systems inherently prioritize one thing over another; and these decisions can affect user beliefs and behaviors. For example, search rankings have been shown to sway voting preferences by more than 20\% among undecided voters~\cite{epstein2015search}, a phenomenon known as the \textit{Search Engine Manipulation Effect}. Although algorithms are designed to optimize relevance and usability, they are also influenced by commercial and political pressures~\cite{introna2000shaping, anderson1997hits}. Despite periodic updates aimed at improving fairness, the inner workings of these systems remain opaque~\cite{bozdag2013bias}, and users are often unaware of how search results are curated.}

\rev{In addition, users bring cognitive biases to search. People often anchor on early results, seek confirmatory evidence, and overlook contradictions, especially in political searches~\cite{azzopardi2021cognitive}. These behaviors increase the impact of ranking, making it critical to study how political cues in queries affect what users see.}

In the present work, we address these questions by comparing the results for politically inclined queries with those for neutral queries across three prominent news search engines: Google News, Bing News, and DuckDuckGo News. These sites were chosen due to their status as the most frequently visited search engines in the United States\footnote{\url{https://www.statista.com/forecasts/997254/most-used-search-engines-by-brand-in-the-us}}. We deliberately exclude Yahoo, as it is generally powered by Bing\footnote{\url{https://techcrunch.com/2015/04/16/microsoft-and-yahoo-renew-search-allian/}}, and focus on news search engine results pages (SERPs), as they often present a distinct framing of topics~\cite{alam2014analyzing}. \rev{Compared to general web search, news verticals prioritize journalistic content and differ systematically in how they rank sources, including emphasis on certain outlets or viewpoints~\cite{alam2014analyzing}. As such, they serve as a critical locus for studying how politically relevant information is curated and framed~\cite{trielli2022algorithmic, urman2022matter}.} Specifically, we aim to address the following research questions:
\begin{enumerate}
    \item \textbf{Semantic Polarity} (\textbf{RQ1}). How do the news headlines returned for politically left- and  right-leaning queries differ in semantic content from those generated by neutral queries across various search engines? 
    \item \textbf{Framing Dynamics} (\textbf{RQ2}). How do search engines differ in their framing of politically salient topics across left-, right-, and neutral-leaning queries? 
    \item \textbf{Source Dynamics} (\textbf{RQ3}). Do search engines disproportionately surface content from a limited number of dominant sources, and how does this tendency align with the political leanings of those sources?
    \item \textbf{Bias and Factuality across Ideologies} (\textbf{RQ4}). How do different search engines prioritize bias and factuality when curating top search results?
\end{enumerate}
\noindent\textbf{Findings.} Through a series of experiments and data analysis detailed in the remainder of this paper, we have found that: (RQ1) Left-leaning and right-leaning queries return more ideologically inclined results. Neutral queries tend to bridge ideological divides. Results from Google News were less likely to reflect the ideology of the query; (RQ2) Bing and DuckDuckGo are more likely to return results having \textit{Episodic}, \textit{Human-interest}, and \textit{Conflict} frames compared to Google, with politically-oriented queries amplifying \textit{Game/Strategy} and \textit{Conflict} frames; (RQ3) search engines surface different sources disproportionately, with Bing and DuckDuckGo more closely aligned than Google; (RQ4) Google provides the most politically-neutral sources, while Bing and DuckDuckGo are more likely to return right-leaning sources; curiously, right-leaning queries unexpectedly shifted toward left-leaning sources on all platforms. Factuality is consistent across search engines, but sources are less factual for political queries.

\section{Data Collection Methodology}

We analyzed search engine results from the \rev{news verticals} of three major platforms: Bing, Google, and DuckDuckGo. \rev{To evaluate political variation in content curation, we used 57 issue-based queries derived from ProCon.org\footnote{\url{https://www.britannica.com/procon}}, a non-partisan source known for curating socially and politically contentious topics~\cite{gezici2021evaluation,procon2023}. To introduce ideological cues, we appended \texttt{Biden} or \texttt{Trump} to each query, following prior work that uses political figure names to simulate partisan framing in search audits~\cite{trielli2022algorithmic,urman2022matter}. Importantly, these are queries about left- or right-leaning political figures, rather than indicators of any user's own ideology.}

\subsection{Search Engine Data}

Collecting search engine data presents unique challenges due to the dynamic and constantly evolving nature of search results. For this study, we utilized the SerpWow API\footnote{\url{https://trajectdata.com/serp/serp-wow-api/}} to systematically gather search engine data within a single ten-hour window in Fall of 2024. 

\paragraph{Mitigating Data Collection Bias} To ensure a fair comparison across platforms, we synchronized the data collection process by retrieving results for each keyword from all three search engines simultaneously. This approach minimized temporal variations and allowed for a consistent evaluation of search engine outputs. Prior work has found no significant differences in search engine data collected within a 10–15 minute range~\cite{gezici2021evaluation}, further supporting the robustness of our synchronized data collection methodology. We specifically set the location to the United States and the language to English, retrieving results from up to the top five pages (if available); otherwise, the collection stopped when no further results were provided for a given query. Furthermore, the SerpWow API uses thousands of distributed proxy services to abate any locality-effects. For analysis, we focus on the top 20 results, as research indicates that users typically gather information from the top 10 results~\cite{schultheiss2021users} and rarely venture past the first 25 results~\cite{glenski2017consumers}.

\begin{table}[t]
\centering
\caption{Number of search results collected for politically right-leaning, left-leaning, and neutral queries across three search engines.}
\label{tab:news_neutral_sentiment}
\begin{tabular}{lccc}
    \toprule
    \textbf{Search Engine} & \textbf{Right} & \textbf{Left} & \textbf{Neutral} \\ 
    \midrule
    Google News            & 3,809          & 3,542         & 4,697            \\ 
    DuckDuckGo News        & 4,749          & 4,900         & 5,825            \\ 
    Bing News              & 506            & 502           & 582              \\ 
    \bottomrule
\end{tabular}
\end{table}

The total number of search results across different search engines for right-leaning, left-leaning, and neutral queries are presented in Table~\ref{tab:news_neutral_sentiment}. 
A $\chi^2$ test of independence was conducted to examine the relationship between search engine and political leaning. The test revealed a statistically significant association, $\chi^2(4, N = 29112) = 18.09, p = 0.0012 $. However, the effect size, measured using Cramér's V, was 0.018, indicating a very weak effect size in the number of results returned for  search engines and political leaning.

% For Google News, the results are 3,809 for right-leaning queries, 3,542 for left-leaning queries, and 4,697 for universal queries. DuckDuckGo News shows 4,749 results for right-leaning, 4,900 for left-leaning, and 5,825 for universal queries. Bing News, on the other hand, returns 506 results for right-leaning, 502 for left-leaning, and 582 for universal queries. These values highlight the distribution of search results across the three engines.

% \section{Framing}
% interested not in what is told but how is it told

\section{Semantic Polarity}

Headlines are the first point of engagement in a news article, shaping the reader’s expectations and framing their interpretation of the content~\cite{konnikova2014headlines}. They convey significance, emphasize gravitas, and reinforce authority~\cite{papacharissi2018importance}. Due to this central role, we focus on headlines as the primary unit of analysis and ask: (\textbf{RQ1}) How do the news headlines returned for politically left- and right-leaning queries differ in semantic content from those generated by neutral queries across various search engines?

We examine the semantic polarity of search engines by comparing how their results for politically inclined queries differ from those for neutral queries. This analysis assesses the impact of political keywords on the content and framing of retrieved headlines. Previous studies have measured semantic polarization by analyzing linguistic differences in broadcast news and social media discourse~\cite{mowshowitz2002assessing, ding2023same}, often using contextualized language embeddings to quantify shifts in framing and context.

However, in search engine rankings individual results are not independent but rather part of an ordered list influenced by underlying ranking algorithms. This order is crucial to how the user views and interprets information. So, to analyze the alignment or divergence between result lists (\eg, comparing a neutral query against a politicized query), it is necessary to perform pairwise comparisons. Pairwise cosine similarity allows us to capture the semantic relationship between each pair of articles across result vectors. 

Specifically, for each neutral query, we define left- and right-leaning variants. When these queries are submitted to a search engine, the engine returns results for each variant. We then used MPNetv2~\cite{reimers-2019-sentence-bert}, a pretrained sentence transformer, to encode each headline into to a 128-dimensional vector space and computed the pairwise comparison for \rev{10,000 random headline-pairs} using the cosine distance.

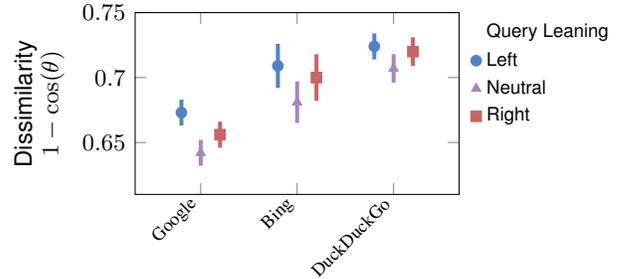
\begin{figure}[t]
    \centering
    %% --- Left Query ---
\pgfplotstableread{
engine mean ci_lower ci_upper
Google_news 0.673 0.662 0.683
Bing_news 0.709 0.691 0.726
Duckduckgo_news 0.724 0.713 0.734
}\datatableleft

%% --- Right Query ---
\pgfplotstableread{
engine mean ci_lower ci_upper
Google_news 0.656 0.646 0.666
Bing_news 0.7 0.682 0.718
Duckduckgo_news 0.72 0.709 0.731
}\datatableright

%% --- Universal Query ---
\pgfplotstableread{
engine mean ci_lower ci_upper
Google_news 0.642 0.633 0.652
Bing_news 0.681 0.665 0.697
Duckduckgo_news 0.707 0.696 0.718
}\datatableuniversal

\begin{tikzpicture}
\begin{axis}[
    width=.7\linewidth,
    height=4cm,
    enlarge x limits=0.2,
    ymin=0.61, ymax=0.75,
    ylabel={Dissimilarity \\ $1-\cos(\theta)$},
    ylabel style={align=center},
    xlabel={},
    %symbolic x coords={Google_news, Bing_news, Duckduckgo_news},
    xtick={0,1,2},
    xticklabels = {Google, Bing, DuckDuckGo},
    xticklabel style={rotate=45, anchor=east, font=\scriptsize},
    legend style={at={(1.01,1.0)}, anchor=north west, legend columns=1},
    legend cell align={left},
    %error bars/.cd,
    %    y dir=both, y explicit,
    %nodes near coords,
    %nodes near coords align={vertical}
]

\addplot+[
    only marks,
    mark=*,
    mark options={fill=left, draw=left},
    mark size=2.0pt,
    forget plot,
    error bars/.cd,
        y dir=both,
        y explicit,
        error mark=bar,
        error bar style={line width=1.5pt, draw=left},
] table[
    x expr=\coordindex - 0.2,
    y=mean,
    y error expr=\thisrow{ci_upper} - \thisrow{mean}
] \datatableleft;

\addlegendimage{only marks, mark=o, fill=none, draw=white}
\addlegendentry{Query Leaning}

\addlegendimage{only marks, mark=*, fill=left, draw=left}
\addlegendentry{Left}

\addplot+[
    only marks,
    mark=triangle*,
    mark options={fill=center, draw=center},
    mark size=2.0pt,
    forget plot,
    error bars/.cd,
        y dir=both,
        y explicit,
        error mark=bar,
        error bar style={line width=1.5pt, draw=center},
] table[
    x expr=\coordindex - 0,
    y=mean,
    y error expr=\thisrow{ci_upper} - \thisrow{mean}
] \datatableuniversal;

\addlegendimage{only marks, mark=triangle*, fill=center, draw=center}
\addlegendentry{Neutral}

\addplot+[
    only marks,
    mark=square*,
    mark options={fill=right, draw=right},
    mark size=2.0pt,
    forget plot,
    error bars/.cd,
        y dir=both,
        y explicit,
        error mark=bar,
        error bar style={line width=1.5pt, draw=right},
] table[
    x expr=\coordindex + 0.2,
    y=mean,
    y error expr=\thisrow{ci_upper} - \thisrow{mean}
] \datatableright;

\addlegendimage{only marks, mark=square*, fill=right, draw=right}
\addlegendentry{Right}

\end{axis}
\end{tikzpicture}
    \caption{Cosine dissimilarity ($1-\cos\theta$) between search result headlines, grouped by query leaning and search engine. Values reflect estimated marginal means from a linear mixed-effects model (LMM) with random intercepts for query key and fixed effects for search engine and query leaning. Error bars represent 95\% confidence intervals. Results indicate that dissimilarity varies by query type and platform, with right-leaning queries often producing more internally divergent results.}
    \label{fig:dissimilarity}
\end{figure}

% Boxplot distributions of these cosine distances are illustrated in Fig.~\ref{fig:dissimilarity}. 

\rev{To account for dependencies across repetitions in the pairwise comparisons, we fit a linear mixed-effects model (LMM) predicting pairwise semantic dissimilarity using cosine distance between headline embeddings. The model included fixed effects for search engine and query leaning, and random intercepts for each headline to control for repeated items. This allowed us to examine how both platform design and political query framing shape the internal consistency of search results. We find that both factors significantly influence headline similarity, with some platforms returning more semantically cohesive results than others, and neutral queries generally producing more consistent content than political ones.}

\rev{Figure~\ref{fig:dissimilarity} presents estimated marginal cosine dissimilarity scores between headlines, grouped by query leaning and search engine. We find that queries with political cues, especially left-leaning ones, tend to generate more divergent headline content, particularly on Bing and DuckDuckGo. By contrast, neutral (\ie, non-political) queries return more semantically similar headlines, suggesting broader alignment in framing. Google shows the most stable dissimilarity patterns across query types, indicating a greater degree of internal coherence in its retrieved results regardless of ideological input.}

%A repeated measures ANOVA was conducted to assess the effect of the pairs of semantic distance across three search engines. The analysis revealed a significant main effect of condition for all three search engines: Google News (\(F(2, 112) = 102.77\)), Bing News (\(F(2, 104) = 236.17\)), and DuckDuckGo News (\(F(2, 100) = 476.96\)), with \(p < 0.001\) in all cases. This indicates that the semantic distances differed significantly across the three conditions for all search engines.

%Post-hoc pairwise comparisons using Tukey's HSD test revealed that the semantic distances for the comparisons neutral vs left-leaning and neutral vs right-leaning queries were statistically more similar than left-leaning vs right-leaning queries, with \(p < 0.001\) for both. These results suggest that the semantic distance between the two political news spheres is much larger than between neutral and political comparisons. 

These results are not unexpected, but serves as a sanity check, demonstrating that the methodology is effectively capturing meaningful semantic differences. The larger semantic distance between left-leaning and right-leaning content, compared to the neutral comparisons, validates the approach's sensitivity to ideological variation in the data. Next, we will look how these differences impact how users see the news from various perspectives.

\section{Framing Dynamics}

%condensed version!
The way information is presented significantly impacts how readers perceive and understand news stories. This concept, known as framing, originates from sociology~\cite{goffman1974frame}, where Entman~\citeyearpar{entman1993framing} describes framing as the process of highlighting specific elements of a perceived reality to promote particular interpretations and evaluations, focusing on ``selection'' and ``salience.'' The crucial questions here are: Who chooses what to highlight, and why does it matter to the audience~\cite{entman1993framing}?
% In this context, salience refers to making information more noticeable, meaningful, or memorable to the audience~\cite{entman1993framing}.

Primarily, journalists decide on news story headlines and ledes, effectively acting as news framers~\cite{bruggemann2014between, reese2010finding, baden2019framing}. However, this is not done in a vacuum; articles undergo extensive editorial processes to meet publishers' standards before public release. This frame-building involves a dynamic interplay between journalists, editors, elites~\cite{gans1979deciding, tuchman1978making}, and social movements~\cite{cooper2002media}.

Framing plays a pivotal role in shaping information and influencing decisions~\cite{kahneman1984choices}. Politicians and journalists often compete to control news narratives and guide public opinion~\cite{riker1986art}.

In this section, we address framing dynamics by asking: (\textbf{RQ2}) How do search engines frame politically salient topics across left-, right-, and neutral-leaning queries?

Framing theory distinguishes between issue-specific frames, which focus on particular topics, and generic frames, which apply across various stories~\cite{de2005news}. Scholars typically classify frames into six archetypes: (1) conflict, (2) game/strategy, (3) thematic/issue, (4) economic consequences, (5) episodic, and (6) human interest~\cite{baden2019framing}. For example, conflict frames emphasize competing viewpoints, while game/strategy frames highlight the tactics behind social and political struggles~\cite{lawrence2000game}. Although there may be several other types of frames and frame-ontologies, prior work has found that these six appear most frequently in news~\cite{de2005news}; for a more thorough exploration of frames see the chapter by~\citet{baden2019framing}.

Our research builds on this framework by categorizing news headlines from our dataset into these specific frames. Through this, we aim to uncover how framing shapes public discourse and influences political perceptions.

\begin{figure}[t]
    \centering
        \begin{tikzpicture}
        \begin{groupplot}[
            group style={group size=2 by 1, horizontal sep=0.15cm},
            height=5.5cm,width=5cm,
            ybar stacked,
            % width=0.4\textwidth,
            % height=6cm,
            ymin=0, ymax=100,
            xlabel = {},
            ylabel = {},
            title style={font=\small},
            x tick label style={rotate=45,anchor=east, font=\scriptsize},
            enlarge x limits=0.4,
            nodes near coords,
            nodes near coords style={
                font=\tiny,
                color=black,
                /pgf/number format/fixed,
                /pgf/number format/precision=1
            },
            every axis plot/.append style={
                line width=0.2pt,
                draw=black!30 % subtle border
            },
            % legend style={
            %     at={(0.8,-0.65)},
            %     anchor=north,
            %     legend columns=3,
            %     cells={anchor=west},
            %     font=\scriptsize,
            %     draw=none,
            %     row sep=2pt,
            %     column sep=10pt
            % },
            ymajorgrids=true,
            grid style={dashed, gray!30}
        ]
        
        % First plot: By Search Engine
        \nextgroupplot[
            symbolic x coords={Google, Bing, DuckDuckGo},
            bar width=14pt,
            xtick=data,
            ylabel={Percentage (\%)},
            xlabel={Search Engine},    
            xlabel style={yshift=6pt},
            legend style={
                at={(1.1,-0.38)},
                anchor=north,
                legend columns=3,
                cells={anchor=west},
                font=\scriptsize,
                draw=none,
                fill=none,
                row sep=2pt,
                column sep=1pt
            },
            legend entries = {
                Conflict,
                Econ. Consequences,
                Episodic,
                Game/Strategy,
                Human Interest,
                Thematic/Issue
            }  
        ]
\addplot+[ybar, draw=conflict, fill=conflict!40] 
    coordinates {(Google,15.41) (Bing,22.77) (DuckDuckGo,23.50)};
\addplot+[ybar, draw=econ, fill=econ!40] 
    coordinates {(Google,7.56) (Bing,7.72) (DuckDuckGo,7.58)};
\addplot+[ybar, draw=episodic, fill=episodic!40] 
    coordinates {(Google,1.12) (Bing,3.38) (DuckDuckGo,4.50)};
\addplot+[ybar, draw=strategy, fill=strategy!40] 
    coordinates {(Google,15.48) (Bing,16.91) (DuckDuckGo,19.10)};
\addplot+[ybar, draw=human, fill=human!40] 
    coordinates {(Google,3.09) (Bing,5.64) (DuckDuckGo,6.50)};
\addplot+[ybar, draw=thematic, fill=thematic!40] 
    coordinates {(Google,57.34) (Bing,43.58) (DuckDuckGo,38.82)};

        % Second plot: By Political Alignment
        \nextgroupplot[
            symbolic x coords={Left, Neutral, Right},
            bar width=14pt,
            xtick=data,
            xlabel={Query Leaning},
            xlabel style={yshift=-7pt},
            ylabel={},
            yticklabels={}
        ]
\addplot+[ybar, draw=conflict, fill=conflict!40] coordinates {(Left,22.64) (Neutral,15.48)  (Right,25.28)};
\addplot+[ybar, draw=econ, fill=econ!40] coordinates { (Left,9.31) (Neutral,7.43) (Right,7.35)};
\addplot+[ybar, draw=episodic, fill=episodic!40] coordinates {(Left,3.37) (Neutral,2.78)  (Right,3.12)};
\addplot+[ybar, draw=strategy, fill=strategy!40] coordinates {(Left,20.58) (Neutral,10.13) (Right,23.38)};
\addplot+[ybar, draw=human, fill=human!40] coordinates {(Left,4.72) (Neutral,5.98)  (Right,5.14)};
\addplot+[ybar, draw=thematic, fill=thematic!40] coordinates {(Left,39.38) (Neutral,58.20) (Right,35.73)};
        
        \end{groupplot}

    \end{tikzpicture}
    \caption{Percentage distribution of news frames across search engines and query leaning. Thematic/Issue frames dominate across all search engines and sources, while Episodic frames are the least common. DuckDuckGo exhibits a more balanced frame distribution, whereas partisan sources (left and right) show higher representation of Conflict and Game/Strategy frames.}
    \label{fig:frame_comparison}
\end{figure}
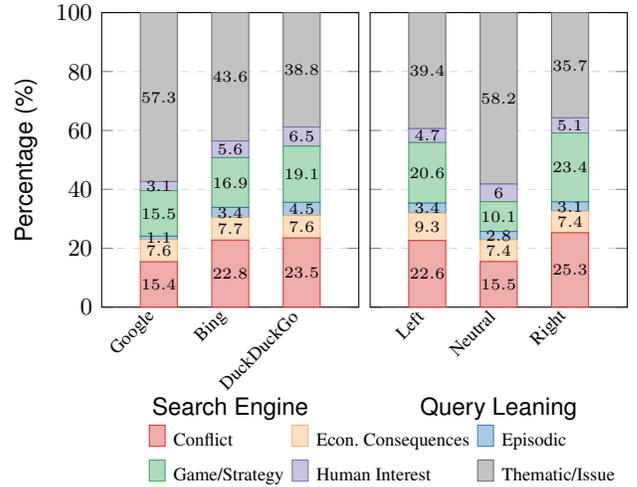

\begin{figure}[t]
\centering
%% --- Conflict Frame ---
\pgfplotstableread{
engine mean ci_lower ci_upper
Google 0.52 0.48 0.55
Bing 0.52 0.48 0.56
Duckduckgo 0.54 0.51 0.56
}\datatableconflict

%% --- Economic Frame ---
\pgfplotstableread{
engine mean ci_lower ci_upper
Google 0.52 0.47 0.57
Bing 0.48 0.41 0.54
Duckduckgo 0.51 0.47 0.56
}\datatableeconomic

%% --- Episodic Frame ---
\pgfplotstableread{
engine mean ci_lower ci_upper
Google 0.40 0.30 0.51
Bing 0.68 0.59 0.77
Duckduckgo 0.59 0.54 0.64
}\datatableepisodic

%% --- Human Frame ---
\pgfplotstableread{
engine mean ci_lower ci_upper
Google 0.53 0.45 0.61
Bing 0.56 0.50 0.63
Duckduckgo 0.58 0.53 0.62
}\datatablehuman

%% --- Strategy Frame ---
\pgfplotstableread{
engine mean ci_lower ci_upper
Google 0.50 0.47 0.54
Bing 0.53 0.49 0.57
Duckduckgo 0.52 0.49 0.55
}\datatablestrategy

%% --- Thematic Frame ---
\pgfplotstableread{
engine mean ci_lower ci_upper
Google 0.54 0.53 0.56
Bing 0.48 0.45 0.50
Duckduckgo 0.47 0.45 0.49
}\datatablethematic

%% --- Conflict Frame by Leaning ---
\pgfplotstableread{
leaning mean ci_lower ci_upper
Left 0.55 0.52 0.58
Universal 0.52 0.48 0.56
Right 0.51 0.49 0.54
}\datatableconflictleaning

%% --- Economic Frame by Leaning ---
\pgfplotstableread{
leaning mean ci_lower ci_upper
Left 0.50 0.45 0.55
Universal 0.52 0.46 0.58
Right 0.51 0.46 0.56
}\datatableeconomicleaning

%% --- Episodic Frame by Leaning ---
\pgfplotstableread{
leaning mean ci_lower ci_upper
Left 0.61 0.54 0.67
Universal 0.66 0.58 0.74
Right 0.50 0.43 0.56
}\datatableepisodicleaning

%% --- Human Frame by Leaning ---
\pgfplotstableread{
leaning mean ci_lower ci_upper
Left 0.52 0.46 0.59
Universal 0.56 0.51 0.61
Right 0.60 0.55 0.65
}\datatablehumanleaning

%% --- Strategy Frame by Leaning ---
\pgfplotstableread{
leaning mean ci_lower ci_upper
Left 0.50 0.47 0.53
Universal 0.52 0.47 0.56
Right 0.53 0.50 0.56
}\datatablestrategyleaning

%% --- Thematic Frame by Leaning ---
\pgfplotstableread{
leaning mean ci_lower ci_upper
Left 0.51 0.49 0.53
Universal 0.51 0.49 0.53
Right 0.49 0.47 0.51
}\datatablethematicleaning

\begin{tikzpicture}
\begin{groupplot}[
    group style={
        group size=1 by 2,
        horizontal sep=0.2cm,
        vertical sep=1.0cm,
    },    
    legend style={
        at={(0.5,-0.25)},
        anchor=north,
        draw=none,
        /tikz/every even column/.style={anchor=west},
        fill=none,
        font=\scriptsize,
    },
    legend columns=-1,
    ymajorgrids=true,
    ymin=0, ymax=1,
    width=0.4\textwidth,
    height=4cm,
    y dir=reverse,
]

\nextgroupplot[
    ytick={0.1,0.5,0.9},
    yticklabel style={font=\scriptsize},
    yticklabels={Top, Mid, Bot},
    xlabel={News Search Engine},
    xlabel style={font=\footnotesize, yshift=5pt},
    xmin=-0, xmax=2,
    enlarge x limits=0.4,
    xtick={0,1,2},
    xticklabels={Google, Bing, DuckDuckGo},
    xticklabel style={font=\scriptsize},
    ylabel={Ranking},
    %legend columns=3
]

\addplot+[
    only marks,
    mark=*,
    mark options={fill=conflict, draw=conflict},
    mark size=2.0pt,
    forget plot,
    error bars/.cd,
        y dir=both,
        y explicit,
        error mark=bar,
        error bar style={line width=1.5pt, draw=conflict},
] table[
    x expr=\coordindex - 0.3,
    y=mean,
    y error expr=\thisrow{ci_upper} - \thisrow{mean}
] \datatableconflict;

% moderate
\addplot+[
    only marks,
    mark=triangle*,
    mark options={fill=econ, draw=econ},
    mark size=2.3pt,
    forget plot,
    error bars/.cd,
        y dir=both,
        y explicit,
        error mark=bar,
        error bar style={line width=1.5pt, draw=econ},
] table[
    x expr=\coordindex - 0.2,
    y=mean,
    y error expr=\thisrow{ci_upper} - \thisrow{mean}
] \datatableeconomic;

\addplot+[
    only marks,
    mark=square*,
    mark options={fill=episodic, draw=episodic},
    mark size=2.0pt,
    forget plot,
    error bars/.cd,
        y dir=both,
        y explicit,
        error mark=bar,
        error bar style={line width=1.5pt, draw=episodic},
] table[
    x expr=\coordindex - 0.1,
    y=mean,
    y error expr=\thisrow{ci_upper} - \thisrow{mean}
] \datatableepisodic;

\addplot+[
    only marks,
    mark=diamond*,
    mark options={fill=strategy, draw=strategy},
    mark size=2.3pt,
    forget plot,
    error bars/.cd,
        y dir=both,
        y explicit,
        error mark=bar,
        error bar style={line width=1.5pt, draw=strategy},
] table[
    x expr=\coordindex ,
    y=mean,
    y error expr=\thisrow{ci_upper} - \thisrow{mean}
] \datatablestrategy;

\addplot+[
    only marks,
    mark=pentagon*,
    mark options={fill=human, draw=human},
    mark size=2.3pt,
    forget plot,
    error bars/.cd,
        y dir=both,
        y explicit,
        error mark=bar,
        error bar style={line width=1.5pt, draw=human},
] table[
    x expr=\coordindex + 0.1,
    y=mean,
    y error expr=\thisrow{ci_upper} - \thisrow{mean}
] \datatablehuman;

\addplot+[
    only marks,
    mark=triangle*,
    mark options={rotate=180, fill=thematic, draw=thematic},
    mark size=2.2pt,
    forget plot,
    error bars/.cd,
        y dir=both,
        y explicit,
        error mark=bar,
        error bar style={line width=1.5pt, draw=thematic},
] table[
    x expr=\coordindex + 0.2,
    y=mean,
    y error expr=\thisrow{ci_upper} - \thisrow{mean}
] \datatablethematic;

\nextgroupplot[
    ytick={0.1,0.5,0.9},
    yticklabel style={font=\scriptsize},
    yticklabels={Top, Mid, Bot},
    xlabel={Query Leaning},
    xlabel style={font=\footnotesize, yshift=5pt},
    xmin=-0, xmax=2,
    enlarge x limits=0.4,
    xtick={0,1,2},
    xticklabels={Left, Neutral, Right},
    xticklabel style={font=\scriptsize},
    ylabel={Ranking},
    %legend columns=3
]

\addplot+[
    only marks,
    mark=*,
    mark options={fill=conflict, draw=conflict},
    mark size=2.0pt,
    forget plot,
    error bars/.cd,
        y dir=both,
        y explicit,
        error mark=bar,
        error bar style={line width=1.5pt, draw=conflict},
] table[
    x expr=\coordindex - 0.3,
    y=mean,
    y error expr=\thisrow{ci_upper} - \thisrow{mean}
] \datatableconflictleaning;

\addlegendimage{only marks, mark=*, mark size=1.5pt, fill=conflict, draw=conflict}
\addlegendentry{Conflict}

% moderate
\addplot+[
    only marks,
    mark=triangle*,
    mark options={fill=econ, draw=econ},
    mark size=2.3pt,
    forget plot,
    error bars/.cd,
        y dir=both,
        y explicit,
        error mark=bar,
        error bar style={line width=1.5pt, draw=econ},
] table[
    x expr=\coordindex - 0.2,
    y=mean,
    y error expr=\thisrow{ci_upper} - \thisrow{mean}
] \datatableeconomicleaning;
\addlegendimage{only marks, mark=triangle*, mark size=2.0pt, fill=econ, draw=econ}
\addlegendentry{Economic}

\addplot+[
    only marks,
    mark=square*,
    mark options={fill=episodic, draw=episodic},
    mark size=2.3pt,
    forget plot,
    error bars/.cd,
        y dir=both,
        y explicit,
        error mark=bar,
        error bar style={line width=1.5pt, draw=episodic},
] table[
    x expr=\coordindex - 0.1,
    y=mean,
    y error expr=\thisrow{ci_upper} - \thisrow{mean}
] \datatableepisodicleaning;
\addlegendimage{only marks, mark=square*, mark size=1.7pt, fill=episodic, draw=episodic}
\addlegendentry{Episodic}

\addplot+[
    only marks,
    mark=diamond*,
    mark options={fill=strategy, draw=strategy},
    mark size=2.3pt,
    forget plot,
    error bars/.cd,
        y dir=both,
        y explicit,
        error mark=bar,
        error bar style={line width=1.5pt, draw=strategy},
] table[
    x expr=\coordindex ,
    y=mean,
    y error expr=\thisrow{ci_upper} - \thisrow{mean}
] \datatablestrategyleaning;
\addlegendimage{only marks, mark=diamond*, mark size=1.7pt, fill=strategy, draw=strategy}
\addlegendentry{Game}

\addplot+[
    only marks,
    mark=pentagon*,
    mark options={fill=human, draw=human},
    mark size=2.3pt,
    forget plot,
    error bars/.cd,
        y dir=both,
        y explicit,
        error mark=bar,
        error bar style={line width=1.5pt, draw=human},
] table[
    x expr=\coordindex + 0.1,
    y=mean,
    y error expr=\thisrow{ci_upper} - \thisrow{mean}
] \datatablehumanleaning;
\addlegendimage{only marks, mark=pentagon*, mark size=1.7pt, fill=human, draw=human}
\addlegendentry{Human}

\addplot+[
    only marks,
    mark=triangle*,
    mark options={rotate=180, fill=thematic, draw=thematic},
    mark size=2.2pt,
    forget plot,
    error bars/.cd,
        y dir=both,
        y explicit,
        error mark=bar,
        error bar style={line width=1.5pt, draw=thematic},
] table[
    x expr=\coordindex + 0.2,
    y=mean,
    y error expr=\thisrow{ci_upper} - \thisrow{mean}
] \datatablethematicleaning;
\addlegendimage{only marks, mark=triangle*,
    rotate=180, fill=thematic, draw=thematic}
\addlegendentry{Thematic}

\end{groupplot}
\end{tikzpicture}
\caption{Tukey post-hoc comparisons of average ranking positions for each frame across search engines. These results highlight how ranking algorithms shape the visibility of narrative types across platforms.}
\label{fig:rq2_tukey}
\end{figure}
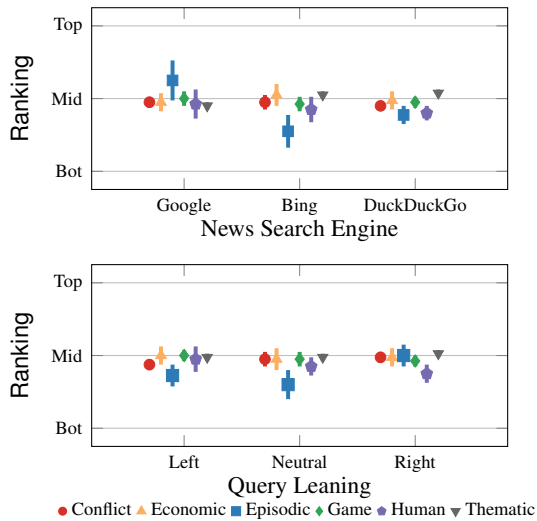

\begin{figure}[t]
\begin{tikzpicture}
\begin{groupplot}[
    group style={
        group name=coefplot,
        group size=1 by 3, % 1 column, 3 rows
        vertical sep=0.15cm, % Vertical separation between plots
    },
    width=.85\linewidth,
    xlabel style = {align=center, font=\footnotesize},
    xlabel={\textcolor{blue}{$\xleftarrow{\textrm{Higher Ranking}}$}\hspace{1.5cm}\textcolor{red}{$\xrightarrow{\textrm{Lower Ranking}}$} \\ Coefficients (Normalized Rank) },
    ylabel={},
    ylabel style = {font=\sffamily\footnotesize},
    yticklabel style = {font=\sffamily\scriptsize},
    xticklabel style = {font=\sffamily\scriptsize},
    xmax = 0.11,
    xmin = -0.11,
    xtick = {-0.1, 0, 0.1},
    scatter/classes={
    conflict={mark=*,red}, 
    economic={mark=*,blue}, 
    episodic={mark=*,teal}, 
    game={mark=*,orange},
    human={mark=*,black}
    },
    scatter,
    clip=false
]

\nextgroupplot[
    %title=First Plot Title,
        ytick={1,2,3,4,5},
        yticklabels={
            Economic,
            Episodic,
            Game,
            Human,
            Thematic
        },
        ylabel={},
        ymin = 0.5,
        ymax = 5.5,
        height=4cm,
        only marks,
        ylabel={Result Frame},
        ylabel style = {align=center, font=\footnotesize},
        xticklabels={},
        xlabel={},      % Suppress x-axis label
]
    % Vertical reference line
    \addplot[dashed, samples=50, smooth, no marks, domain=0:10, black] coordinates {(0,0.5)(0,5.5)};
    %\node[anchor=south east, font=\footnotesize\sffamily] at (axis cs:0.6,2.5) {*** $p<0.01$, ** $p<0.05$, * $p<0.1$};

    \node[black] (baseline) at (axis cs:-0.05,2.1) {\scriptsize{Conflict}};
    \draw[->, bend right=30, thick] (baseline) to[out=-90, in=180] (axis cs:-0.001,1.8);

    \node[anchor=south east, font=\footnotesize\sffamily] at (axis cs:0.1,5.5) {*** $p<0.01$, ** $p<0.05$, * $p<0.1$};

    % Error bars with y-shift
    \addplot [
        scatter src=explicit symbolic,
        visualization depends on={value \thisrow{lab} \as \Label},
        nodes near coords*={\tiny{\sffamily{\Label}}},
        every node near coord/.append style={
            anchor=north, % Align the text to the west of the node
            yshift =1pt
        },
        error bars/.cd, 
        x dir = both, 
        x explicit, 
        error bar style={thick, solid, black},
        error mark options={line width=0.5pt, mark size=2pt, rotate=90}
    ]    
    table[meta=class, x=x, y=y, x error=ex]{
        y x         ex          class   lab     
        1 -0.022   0.018     d {}
        2 0.058   0.025     d *
        3 -0.012   0.013     d {}
        4 0.036   0.019     d {}
        5 -0.030   0.011     d **
};

\nextgroupplot[
    %title=First Plot Title,
        ytick={1,2},
        yticklabels={
            Bing,
            DuckDuckGo
        },
        ylabel={},
        ymin = 0.5,
        ymax = 2.5,
        height=3cm,
        only marks,
        ylabel={Search\\Engine},
        ylabel style = {align=center, font=\footnotesize},
        xticklabels={},
        xlabel={},      % Suppress x-axis label
]
    % Vertical reference line
    \addplot[dashed, samples=50, smooth, no marks, domain=0:10, black] coordinates {(0,0.5)(0,2.5)};
    %\node[anchor=south east, font=\footnotesize\sffamily] at (axis cs:0.6,2.5) {*** $p<0.01$, ** $p<0.05$, * $p<0.1$};

    \node[black] (baseline) at (axis cs:-0.05,1.7) {\scriptsize{Google}};
    \draw[->, bend right=30, thick] (baseline) to[out=-90, in=180] (axis cs:-0.001,1.6);

    % Error bars with y-shift
    \addplot [
        scatter src=explicit symbolic,
        visualization depends on={value \thisrow{lab} \as \Label},
        nodes near coords*={\tiny{\sffamily{\Label}}},
        every node near coord/.append style={
            anchor=north, % Align the text to the west of the node
            yshift =1pt
        },
        error bars/.cd, 
        x dir = both, 
        x explicit, 
        error bar style={thick, solid, black},
        error mark options={line width=0.5pt, mark size=2pt, rotate=90}
    ]    
    table[meta=class, x=x, y=y, x error=ex]{
        y x         ex          class   lab     
        1 -0.027   0.011     d *
        2 -0.025   0.009     d **
};

\nextgroupplot[
        ytick={3,4},
        yticklabels={
            Left, Right
        },
        ylabel={Query \\ Leaning},
        ylabel style = {align=center, font=\footnotesize},
        ymin = 2.5,
        ymax = 4.5,  
        height=3cm,
        only marks,
        title={},
]
    % Vertical reference line
    \addplot[dashed, samples=50, smooth, no marks, domain=0:10, black] coordinates {(0,2.5)(0,4.5)};

    \node[black] (baseline) at (axis cs:0.05,4.0) {\scriptsize{Neutral}};
    \draw[->, bend right=30, thick] (baseline) to[out=90, in=180] (axis cs:0.001,3.8);

    % Error bars with y-shift
    \addplot [        
        scatter src=explicit symbolic,
        visualization depends on={value \thisrow{lab} \as \Label},
        nodes near coords*={\tiny{\sffamily{\Label}}},
        every node near coord/.append style={
            anchor=north, % Align the text to the west of the node
            yshift =1pt
        },
        error bars/.cd, 
        x dir = both, 
        x explicit, 
        error bar style={thick, solid, black},
        error mark options={line width=0.5pt, mark size=2pt, rotate=90}
    ]
    table[meta=class, x=x, y=y, x error=ex]{
        y x         ex          class   lab        
        3 0.001    0.010     d {}
        4 -0.011    0.010     d {}    
        };
    \end{groupplot}
\end{tikzpicture}
\caption{Estimated coefficients from a linear mixed-effects model predicting normalized rank of news results, with random intercepts by query. Episodic frames are ranked lower, while thematic frames are elevated across platforms.  Query leaning has limited effect on result prominence. Error bars represent 95\% confidence intervals.}
\label{fig:coefplot}
\end{figure}
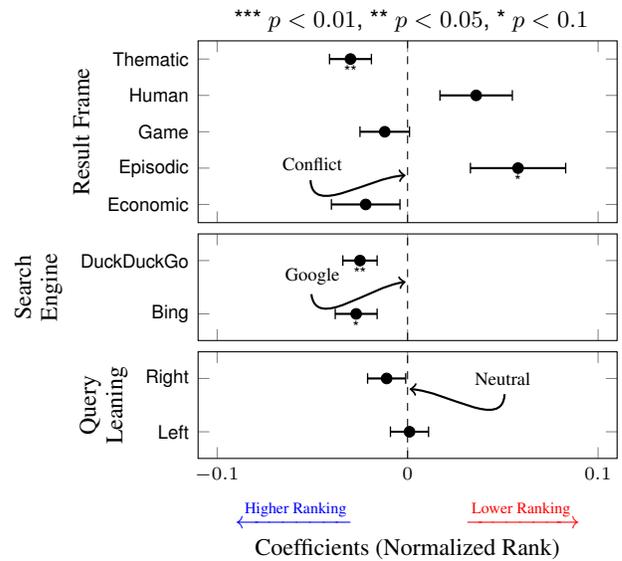

 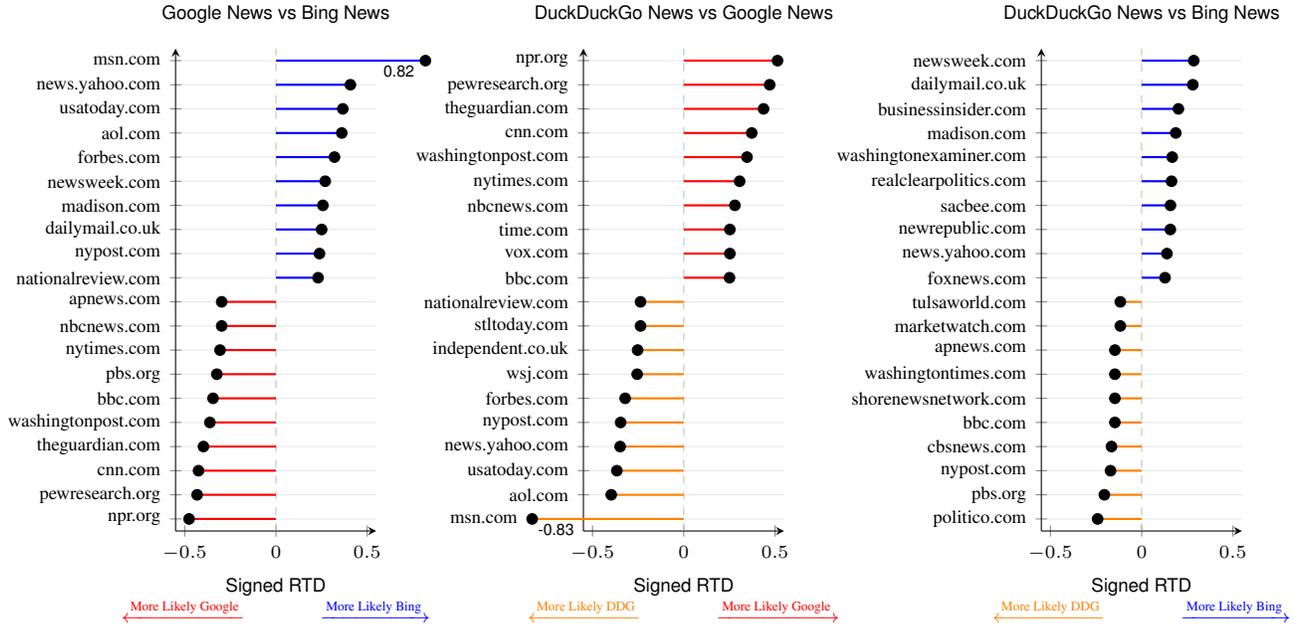
\begin{figure*}[t]
\begin{minipage}{.3\linewidth}
\begin{flushright}
    \pgfplotstableread[col sep=comma]{
Domain,RTD
npr.org,-0.475965291005281
pewresearch.org,-0.43189343962321824
cnn.com,-0.42407194297962597
theguardian.com,-0.3972069424054273
washingtonpost.com,-0.36186650610072774
bbc.com,-0.3446707024348451
pbs.org,-0.3238667641611861
nytimes.com,-0.30610723220108654
nbcnews.com,-0.29719681130465625
apnews.com,-0.29719681130465625
nationalreview.com,0.23095696582380407
nypost.com,0.23861675907757593
dailymail.co.uk,0.2503832550713331
madison.com,0.2574701255283795
newsweek.com,0.2702391010383098
forbes.com,0.3210907858816603
aol.com,0.3608919944299728
usatoday.com,0.3667035672297526
news.yahoo.com,0.4081717356715465
msn.com,0.8195460190905616
}\datatable

\begin{tikzpicture}
\begin{axis}[
    axis lines=left,
    width=4.25cm,
    height=8cm,
    xlabel={},
    ytick=data,
    title={Google News vs Bing News},
    title style={font={\sffamily\scriptsize}},
    % y dir=reverse,
    % yticklabels from table={\datatable}{Domain}
    ytick = {0,...,20},
    xlabel style = {align=center, font=\sffamily\scriptsize},
    xlabel={Signed RTD \\ \textcolor{red}{$\xleftarrow{\textrm{More Likely Google}}$}\hspace{1cm}\textcolor{blue}{$\xrightarrow{\textrm{More Likely Bing}}$}},
    ylabel={},
    yticklabels from table={\datatable}{Domain},
    enlarge y limits={abs=0.5},
    xmin=-0.55,
    xmax=0.55,
    ymajorgrids,
    grid style={line width=.1pt, draw=gray!20},
    yticklabel style={font=\scriptsize, black, align=right},
    xticklabel style = {font=\sffamily\scriptsize},
    clip=false,
    scaled x ticks=false, 
]

% Draw the reference line at x=0
\draw[gray!50, dashed] (axis cs:0,-0.5) -- (axis cs:0,19.5);

% Draw each horizontal line separately
% Draw horizontal bars with black dots at the end
\addplot[red, thick] coordinates {(0,0) (-0.475965291005281,0)};
\addplot[only marks, mark=*, mark size=2pt, black] coordinates {(-0.475965291005281,0)};

\addplot[red, thick] coordinates {(0,1) (-0.43189343962321824,1)};
\addplot[only marks, mark=*, mark size=2pt, black] coordinates {(-0.43189343962321824,1)};

\addplot[red, thick] coordinates {(0,2) (-0.42407194297962597,2)};
\addplot[only marks, mark=*, mark size=2pt, black] coordinates {(-0.42407194297962597,2)};

\addplot[red, thick] coordinates {(0,3) (-0.3972069424054273,3)};
\addplot[only marks, mark=*, mark size=2pt, black] coordinates {(-0.3972069424054273,3)};

\addplot[red, thick] coordinates {(0,4) (-0.36186650610072774,4)};
\addplot[only marks, mark=*, mark size=2pt, black] coordinates {(-0.36186650610072774,4)};

\addplot[red, thick] coordinates {(0,5) (-0.3446707024348451,5)};
\addplot[only marks, mark=*, mark size=2pt, black] coordinates {(-0.3446707024348451,5)};

\addplot[red, thick] coordinates {(0,6) (-0.3238667641611861,6)};
\addplot[only marks, mark=*, mark size=2pt, black] coordinates {(-0.3238667641611861,6)};

\addplot[red, thick] coordinates {(0,7) (-0.30610723220108654,7)};
\addplot[only marks, mark=*, mark size=2pt, black] coordinates {(-0.30610723220108654,7)};

\addplot[red, thick] coordinates {(0,8) (-0.29719681130465625,8)};
\addplot[only marks, mark=*, mark size=2pt, black] coordinates {(-0.29719681130465625,8)};

\addplot[red, thick] coordinates {(0,9) (-0.29719681130465625,9)};
\addplot[only marks, mark=*, mark size=2pt, black] coordinates {(-0.29719681130465625,9)};

\addplot[blue, thick] coordinates {(0,10) (0.23095696582380407,10)};
\addplot[only marks, mark=*, mark size=2pt, black] coordinates {(0.23095696582380407,10)};

\addplot[blue, thick] coordinates {(0,11) (0.23861675907757593,11)};
\addplot[only marks, mark=*, mark size=2pt, black] coordinates {(0.23861675907757593,11)};

\addplot[blue, thick] coordinates {(0,12) (0.2503832550713331,12)};
\addplot[only marks, mark=*, mark size=2pt, black] coordinates {(0.2503832550713331,12)};

\addplot[blue, thick] coordinates {(0,13) (0.2574701255283795,13)};
\addplot[only marks, mark=*, mark size=2pt, black] coordinates {(0.2574701255283795,13)};

\addplot[blue, thick] coordinates {(0,14) (0.2702391010383098,14)};
\addplot[only marks, mark=*, mark size=2pt, black] coordinates {(0.2702391010383098,14)};

\addplot[blue, thick] coordinates {(0,15) (0.3210907858816603,15)};
\addplot[only marks, mark=*, mark size=2pt, black] coordinates {(0.3210907858816603,15)};

\addplot[blue, thick] coordinates {(0,16) (0.3608919944299728,16)};
\addplot[only marks, mark=*, mark size=2pt, black] coordinates {(0.3608919944299728,16)};

\addplot[blue, thick] coordinates {(0,17) (0.3667035672297526,17)};
\addplot[only marks, mark=*, mark size=2pt, black] coordinates {(0.3667035672297526,17)};

\addplot[blue, thick] coordinates {(0,18) (0.4081717356715465,18)};
\addplot[only marks, mark=*, mark size=2pt, black] coordinates {(0.4081717356715465,18)};

\addplot[blue, thick] coordinates {(0,19) (0.8195460190905616,19)};
\addplot[only marks, mark=*, mark size=2pt, black] coordinates {(0.8195460190905616,19)};
\node[anchor=north east, font=\tiny\sffamily] at (axis cs:.81,19.2) {0.82};
% Add the points
% \addplot[
%     only marks,
%     mark=*,
%     mark size=2pt,
%     black
% ] table [y expr=\coordindex] {\datatable};

\end{axis}
\end{tikzpicture}
\end{flushright}
\end{minipage}
\begin{minipage}{.31\linewidth}
\begin{flushright}
    \begin{tikzpicture}
\begin{axis}[
    axis lines=left,
    width=4.25cm,
    height=8cm,
    title={DuckDuckGo News vs Google News},
    title style={font={\sffamily\scriptsize}},
    xlabel={},
    xlabel style = {align=center, font=\sffamily\scriptsize},
    xlabel={Signed RTD \\ \textcolor{orange}{$\xleftarrow{\textrm{More Likely DDG}}$}\hspace{1cm}\textcolor{red}{$\xrightarrow{\textrm{More Likely Google}}$}},
    ytick = {0,...,20},
    xlabel style = {align=center},
    ylabel={},
    yticklabels={
        msn.com~~~~~~~~~~~, aol.com, usatoday.com, news.yahoo.com, nypost.com, forbes.com, wsj.com, independent.co.uk, stltoday.com, nationalreview.com,
        bbc.com, vox.com, time.com, nbcnews.com, nytimes.com, washingtonpost.com, cnn.com, theguardian.com, pewresearch.org, npr.org
    },
    enlarge y limits={abs=0.5},
    xmin=-0.55,
    xmax=0.55,
    ymajorgrids,
    grid style={line width=.1pt, draw=gray!20},
    yticklabel style={font=\scriptsize, black, align=right},
    xticklabel style = {font=\sffamily\scriptsize},
    clip=false,
    scaled x ticks=false,
]

% Draw the reference line at x=0
\draw[gray!50, dashed] (axis cs:0,-0.5) -- (axis cs:0,19.5);

% Bottom 10 (negative values)
\addplot[orange, thick] coordinates {(0,0) (-0.8309633145495144,0)};
\addplot[only marks, mark=*, mark size=2pt, black] coordinates {(-0.8309633145495144,0)};
\node[anchor=north west, font=\tiny\sffamily] at (axis cs:-.85,0.2) {-0.83};

\addplot[orange, thick] coordinates {(0,1) (-0.39732484475944224,1)};
\addplot[only marks, mark=*, mark size=2pt, black] coordinates {(-0.39732484475944224,1)};

\addplot[orange, thick] coordinates {(0,2) (-0.3667035672297526,2)};
\addplot[only marks, mark=*, mark size=2pt, black] coordinates {(-0.3667035672297526,2)};

\addplot[orange, thick] coordinates {(0,3) (-0.3486452357430836,3)};
\addplot[only marks, mark=*, mark size=2pt, black] coordinates {(-0.3486452357430836,3)};

\addplot[orange, thick] coordinates {(0,4) (-0.3458977399802734,4)};
\addplot[only marks, mark=*, mark size=2pt, black] coordinates {(-0.3458977399802734,4)};

\addplot[orange, thick] coordinates {(0,5) (-0.3210907858816603,5)};
\addplot[only marks, mark=*, mark size=2pt, black] coordinates {(-0.3210907858816603,5)};

\addplot[orange, thick] coordinates {(0,6) (-0.2553023804088292,6)};
\addplot[only marks, mark=*, mark size=2pt, black] coordinates {(-0.2553023804088292,6)};

\addplot[orange, thick] coordinates {(0,7) (-0.2528375641103705,7)};
\addplot[only marks, mark=*, mark size=2pt, black] coordinates {(-0.2528375641103705,7)};

\addplot[orange, thick] coordinates {(0,8) (-0.2369286770699933,8)};
\addplot[only marks, mark=*, mark size=2pt, black] coordinates {(-0.2369286770699933,8)};

\addplot[orange, thick] coordinates {(0,9) (-0.2369286770699933,9)};
\addplot[only marks, mark=*, mark size=2pt, black] coordinates {(-0.2369286770699933,9)};

% Top 10 (positive values)
\addplot[red, thick] coordinates {(0,10) (0.25061250202904545,10)};
\addplot[only marks, mark=*, mark size=2pt, black] coordinates {(0.25061250202904545,10)};

\addplot[red, thick] coordinates {(0,11) (0.2526350041594953,11)};
\addplot[only marks, mark=*, mark size=2pt, black] coordinates {(0.2526350041594953,11)};

\addplot[red, thick] coordinates {(0,12) (0.2526350041594953,12)};
\addplot[only marks, mark=*, mark size=2pt, black] coordinates {(0.2526350041594953,12)};

\addplot[red, thick] coordinates {(0,13) (0.2802470663207291,13)};
\addplot[only marks, mark=*, mark size=2pt, black] coordinates {(0.2802470663207291,13)};

\addplot[red, thick] coordinates {(0,14) (0.30610723220108654,14)};
\addplot[only marks, mark=*, mark size=2pt, black] coordinates {(0.30610723220108654,14)};

\addplot[red, thick] coordinates {(0,15) (0.34650759012540294,15)};
\addplot[only marks, mark=*, mark size=2pt, black] coordinates {(0.34650759012540294,15)};

\addplot[red, thick] coordinates {(0,16) (0.37255230886623486,16)};
\addplot[only marks, mark=*, mark size=2pt, black] coordinates {(0.37255230886623486,16)};

\addplot[red, thick] coordinates {(0,17) (0.43730157800205227,17)};
\addplot[only marks, mark=*, mark size=2pt, black] coordinates {(0.43730157800205227,17)};

\addplot[red, thick] coordinates {(0,18) (0.47094923310329995,18)};
\addplot[only marks, mark=*, mark size=2pt, black] coordinates {(0.47094923310329995,18)};

\addplot[red, thick] coordinates {(0,19) (0.5138414677542942,19)};
\addplot[only marks, mark=*, mark size=2pt, black] coordinates {(0.5138414677542942,19)};

\end{axis}
\end{tikzpicture}
\end{flushright}    
\end{minipage}
\begin{minipage}{.3\linewidth}
\begin{flushright}
    \pgfplotstableread[col sep=comma]{
Domain,RTD
politico.com,-0.24144589643440703
pbs.org,-0.2046551487685274
nypost.com,-0.17085500328983935
cbsnews.com,-0.165546614035254
bbc.com,-0.14678274060197732
shorenewsnetwork.com,-0.14678274060197732
washingtontimes.com,-0.14678274060197732
apnews.com,-0.14678274060197732
marketwatch.com,-0.11705913338680028
tulsaworld.com,-0.11705913338680028
foxnews.com,0.1279479071267422
news.yahoo.com,0.1385219302032613
newrepublic.com,0.157075905237913
sacbee.com,0.15773840896121727
realclearpolitics.com,0.16366456391737152
washingtonexaminer.com,0.16738897587911838
madison.com,0.18688165313688446
businessinsider.com,0.20118374876590372
dailymail.co.uk,0.2802205033635435
newsweek.com,0.28545485893509365
}\datatable
\begin{tikzpicture}
\begin{axis}[
    axis lines=left,
    width=4.25cm,
    height=8cm,
    title={DuckDuckGo News vs Bing News},
    title style={font={\sffamily\scriptsize}},
    xlabel={},
    ytick = {0,...,20},
    xlabel style = {align=center, font=\sffamily\scriptsize},
    xlabel={Signed RTD \\ \textcolor{orange}{$\xleftarrow{\textrm{More Likely DDG}}$}\hspace{1cm}\textcolor{blue}{$\xrightarrow{\textrm{More Likely Bing}}$}},
    ylabel={},
    yticklabels from table={\datatable}{Domain},
    enlarge y limits={abs=0.5},
    xmin=-0.55,
    xmax=0.55,
    ymajorgrids,
    grid style={line width=.1pt, draw=gray!20},
    yticklabel style={font=\scriptsize, black, align=right},
    xticklabel style = {font=\sffamily\scriptsize},
    clip=true,
    scaled x ticks=false, 
]

% Draw the reference line at x=0
\draw[gray!50, dashed] (axis cs:0,-0.5) -- (axis cs:0,19.5);

% Bar 11
\addplot[orange, thick] coordinates {(0,0) (-0.24144589643440703,0)};
\addplot[only marks, mark=*, mark size=2pt, black] coordinates {(-0.24144589643440703,0)};

% Bar 12
\addplot[orange, thick] coordinates {(0,1) (-0.2046551487685274,1)};
\addplot[only marks, mark=*, mark size=2pt, black] coordinates {(-0.2046551487685274,1)};

% Bar 13
\addplot[orange, thick] coordinates {(0,2) (-0.17085500328983935,2)};
\addplot[only marks, mark=*, mark size=2pt, black] coordinates {(-0.17085500328983935,2)};

% Bar 14
\addplot[orange, thick] coordinates {(0,3) (-0.165546614035254,3)};
\addplot[only marks, mark=*, mark size=2pt, black] coordinates {(-0.165546614035254,3)};

% Bar 15
\addplot[orange, thick] coordinates {(0,4) (-0.14678274060197732,4)};
\addplot[only marks, mark=*, mark size=2pt, black] coordinates {(-0.14678274060197732,4)};

% Bar 16
\addplot[orange, thick] coordinates {(0,5) (-0.14678274060197732,5)};
\addplot[only marks, mark=*, mark size=2pt, black] coordinates {(-0.14678274060197732,5)};

% Bar 17
\addplot[orange, thick] coordinates {(0,6) (-0.14678274060197732,6)};
\addplot[only marks, mark=*, mark size=2pt, black] coordinates {(-0.14678274060197732,6)};

% Bar 18
\addplot[orange, thick] coordinates {(0,7) (-0.14678274060197732,7)};
\addplot[only marks, mark=*, mark size=2pt, black] coordinates {(-0.14678274060197732,7)};

% Bar 19
\addplot[orange, thick] coordinates {(0,8) (-0.11705913338680028,8)};
\addplot[only marks, mark=*, mark size=2pt, black] coordinates {(-0.11705913338680028,8)};

% Bar 20
\addplot[orange, thick] coordinates {(0,9) (-0.11705913338680028,9)};
\addplot[only marks, mark=*, mark size=2pt, black] coordinates {(-0.11705913338680028,9)};

% Bar 1
\addplot[blue, thick] coordinates {(0,19) (0.28545485893509365,19)};
\addplot[only marks, mark=*, mark size=2pt, black] coordinates {(0.28545485893509365,19)};

% Bar 2
\addplot[blue, thick] coordinates {(0,18) (0.2802205033635435,18)};
\addplot[only marks, mark=*, mark size=2pt, black] coordinates {(0.2802205033635435,18)};

% Bar 3
\addplot[blue, thick] coordinates {(0,17) (0.20118374876590372,17)};
\addplot[only marks, mark=*, mark size=2pt, black] coordinates {(0.20118374876590372,17)};

% Bar 4
\addplot[blue, thick] coordinates {(0,16) (0.18688165313688446,16)};
\addplot[only marks, mark=*, mark size=2pt, black] coordinates {(0.18688165313688446,16)};

% Bar 5
\addplot[blue, thick] coordinates {(0,15) (0.16738897587911838,15)};
\addplot[only marks, mark=*, mark size=2pt, black] coordinates {(0.16738897587911838,15)};

% Bar 6
\addplot[blue, thick] coordinates {(0,14) (0.16366456391737152,14)};
\addplot[only marks, mark=*, mark size=2pt, black] coordinates {(0.16366456391737152,14)};

% Bar 7
\addplot[blue, thick] coordinates {(0,13) (0.15773840896121727,13)};
\addplot[only marks, mark=*, mark size=2pt, black] coordinates {(0.15773840896121727,13)};

% Bar 8
\addplot[blue, thick] coordinates {(0,12) (0.157075905237913,12)};
\addplot[only marks, mark=*, mark size=2pt, black] coordinates {(0.157075905237913,12)};

% Bar 9
\addplot[blue, thick] coordinates {(0,11) (0.1385219302032613,11)};
\addplot[only marks, mark=*, mark size=2pt, black] coordinates {(0.1385219302032613,11)};

% Bar 10
\addplot[blue, thick] coordinates {(0,10) (0.1279479071267422,10)};
\addplot[only marks, mark=*, mark size=2pt, black] coordinates {(0.1279479071267422,10)};

\end{axis}
\end{tikzpicture}
\end{flushright}    
\end{minipage}
 \vspace{-.9cm}
    \caption{Divergence of sources across search engines. The figure illustrates the top 10 most divergent sources based on Rank Turbulence Divergence (RTD) for Google News, Bing News, and DuckDuckGo News. Notable differences in source prioritization are observed, with Google favoring NPR, Pew Research, CNN, and The Guardian, while Bing shows a strong preference for MSN. A similar pattern is seen when comparing Google to DuckDuckGo, while Bing and DuckDuckGo display less divergence, with subtle differences in surfacing sources.}
    \label{fig:sourcertds}
\end{figure*}

\paragraph{Methodology} To classify headlines into different frames, we employed GPT-4 as a labeling tool. Using a carefully designed prompt (provided in the appendix for reference), the model categorized headlines into one of five predefined frames. This approach enabled consistent and scalable labeling of a large dataset while preserving contextual nuance. 

\rev{To assess stability, we examined all headlines that appeared multiple times. GPT-4 produced identical frame labels in 93.75\% of these cases (1,334 out of 1,423), demonstrating high internal consistency. We also conducted a manual validation study: human–human agreement was modest (Cohen’s $\kappa = 0.23$), reflecting the subjective difficulty of the task. Notably, in cases where human coders agreed, GPT-4 matched the majority label with substantial agreement (Cohen’s $\kappa = 0.69$, macro-F$_1 = 0.74$). Like previous work~\cite{ding2022gpt} shows, these results also suggest that GPT-4 may be more consistent and reliable than human annotators in this framing context.}

% data$frames <- relevel(data$frames, ref = "Thematic/Issue ") -- baseline
% data$leanings <- relevel(data$leanings, ref = "universal") -- baseline
% data$search_engine <- relevel(data$search_engine, ref = "google_news") -- baseline

\subsection{Results} 
\rev{Before turning to LMM results, we begin with a descriptive overview of frame distributions across platforms and query leanings. As shown in Fig.~\ref{fig:frame_comparison}, Thematic/Issue frames dominate across all platforms, with DuckDuckGo exhibiting a more balanced distribution compared to Google and Bing. Conflict and Game/Strategy frames are more prominent in partisan sources and right-leaning queries, while Episodic frames are the least frequent overall.} A one-way ANOVA further reveals that average ranking position varies significantly by frame type ($F(5, 7895) = 3.54$, $p = .003$), indicating that certain frames tend to appear higher in search results regardless of political leaning. %This structural prioritization underscores the need to examine how such frames are distributed across ideological query types.

\rev{To assess how framing prominence varies across political query leaning and search engine, we conducted two two-way ANOVAs using ranking position as the dependent variable. Then, because multiple results are returned for each query, we again fit linear mixed-effects models (LMMs) with random intercepts for query and fixed effects for frame type, search engine, query leaning, and ranking. This combination of analyses allows us to identify overall patterns while accounting for within-query dependencies. We also tested models with interaction terms, but these did not significantly improve model fit compared to the fixed-effects-only specification.}

\paragraph{Frame vs Search Engine.}
\rev{We first examined whether search engines differ in how they prioritize different news frames. A two-way ANOVA revealed significant main effects of both frame type ($F(5, 8143) = 6.02$, $p < .001$) and search engine ($F(2, 8143) = 466.66$, $p < .001$), along with a significant interaction between the two ($F(10, 8143) = 5.64$, $p < .001$).} 

\rev{As shown in Fig.~\ref{fig:rq2_tukey} (top), post-hoc Tukey comparisons reveal that Google systematically ranks Episodic frames higher (\ie, earlier in the results), while DuckDuckGo and Bing return Episodic, and Human Interest frames in lower positions. These patterns suggest that search engines structurally shape the narrative emphasis users see, not just through content selection, but also through positional ranking.}

\paragraph{Frame vs Query Leaning.}
\rev{We then tested whether the political orientation of the query influenced the ranking of frames (Fig.~\ref{fig:rq2_tukey} (bottom)). A two-way ANOVA with frame and query leaning as predictors of normalized rank revealed a significant main effect of frame ($F(5, 5107) = 4.42$, $p < .001$), but no main effect of political leaning ($F(2, 5107) = 0.53$, $p = .59$). A modest but significant interaction between frame and leaning ($F(10, 5107) = 1.90$, $p = .041$) suggests that while overall frame rankings remain relatively stable, there are some subtle differences in frame prominence depending on the ideological slant of the query.}

\rev{Together, these findings demonstrate that framing is not only unevenly distributed across platforms and political query types, but also systematically structured in terms of visibility and rank.}

\paragraph{Mixed Effects Model.}

To account for within-query dependencies, we fit an LMM predicting the normalized rank of search results from frame type, search engine, and query leaning, with random intercepts for each query key. Results are visualized in Fig.~\ref{fig:coefplot}.

\rev{We find that frame type significantly influences result prominence. Compared to conflict frames, episodic frames are ranked lower, while thematic frames appear higher in the result list. This suggests that news search engines may elevate more issue-focused content and downrank personal or anecdotal narratives. }

\rev{Query leaning had limited effect. There were no statistically significant differences in ranking between politically neutral queries and those framed as left- or right-leaning, suggesting that ideological slant in the query does not systematically affect where results appear in the ranking.}

\rev{Together, these findings suggest that differences in frame visibility are more strongly shaped by platform architecture and narrative style than by query ideology. Google appears more consistent in its ranking behavior, while Bing and DuckDuckGo elevate more variable frame types.}

\section{Source Dynamics}
% \ap{C6: It would be helpful to better explain what RTD is (in an Appendix possibly so
% as not to eat space) and exactly how you used it. I was not sure what
% it was, or how you used it, after reading the short explanation. Maybe
% a concrete example would help. So, the signed RTD for MSN for Bing vs
% Google is 0.82….what does 0.82 mean? Is this on a scale of 0 to 1 etc?
% Give some intuition. -- this is fixed i believe for the most part}

Search engines are often seen providing broad access to information. However, the challenge lies in determining which data to prioritize for each query~\cite{introna2000shaping}. If only ten news articles can be returned for a query, which ten sources should be chosen? This challenge is amplified by the influence of special interests~\cite{mcchesney1998making}, raising concerns about whether certain sources dominate search results, potentially marginalizing diverse or alternative viewpoints.

To understand this better, in this section we investigate source dynamics in search engines. Specifically, as we ask: (\textbf{RQ3}) Do search engines disproportionately surface content from a limited number of dominant sources, and how does this align with the political leanings of those sources?

\paragraph{Methodology} To examine how source rankings differ across search engines, we use Rank Turbulence Divergence (RTD)~\cite{dodds2023allotaxonometry}, a method that quantifies and visualizes variations in rankings, helping us understand whether alternative media is adequately represented.

We analyze the sources from the search engine results pages using traditional token analysis to determine any relative difference. While various statistical methods can compare these distributions~\cite{cha2007comprehensive,deza2006dictionary}, they often struggle with the Zipfian nature typical of text datasets~\cite{gerlach2016similarity,dodds2023allotaxonometry}. To tackle this, we employ Rank Turbulence Divergence (RTD)~\cite{dodds2023allotaxonometry}, which quantifies the disparity in frequency distributions of sources across different search engines, offering a clearer picture of information diversity in search results. \rev{See the Appendix for details on RTD.}

\iffalse
RTD assesses how elements (\eg, words or tokens) are distributed across the ranks in two lists. The process begins by calculating the divergence for each element based on how its rank differs between the two lists. A control parameter, \( \alpha \), adjusts the emphasis placed on rank differences. For example, smaller values of \( \alpha \) highlight discrepancies in higher-ranked (more frequent) elements, while larger values spread the focus more evenly across all ranks. In this study, we use \( \alpha = \frac{1}{3} \), which provides a balanced evaluation of discrepancies across both common and rare elements, as supported by prior research~\cite{dodds2023allotaxonometry}. The overall RTD score is obtained by summing the divergence values for all elements in the combined lists and normalizing the result \ap{to a range of 0 to 1, where 0 indicates identical rank distributions and 1 represents maximal divergence (no overlap in rankings)}. This score provides a single value that quantifies the degree of discrepancy between the two ranked lists, enabling meaningful comparisons.\ap{See Appendix for mathematical details.}
\fi
 
\begin{table}[t]
\centering
\caption{Rank Turbulence Divergence (RTD) between Source Distributions Across Search Engines}
\vspace{-.3cm}
\label{tab:rtd_values}
% \begin{tabular}{@{}ll@{}}
\begin{tabular}{lll}
\toprule
\multicolumn{2}{l}{\textbf{Search Engine Comparison}}         & \textbf{RTD Value} \\ \midrule
Google News & Bing News & $0.5896$ \\
Google News & DuckDuckGo News & $0.5781$ \\
Bing News & DuckDuckGo News   & $0.3141$ \\ \bottomrule
\end{tabular}
\label{tab:rtd}
\end{table}

\paragraph{Results} Table~\ref{tab:rtd} presents the RTD values between the source distributions across different search engines. Statistical tests were conducted by performing 1000 bootstrap iterations of 100 samples each. 95\% confidence intervals are on the order of $10^{-16}$ and are therefore omitted from the table. The RTD value between Bing and DuckDuckGo suggests a closer alignment in the types of sources these engines prioritize compared to Google. 

Although these results are interesting, we are mostly interested in which sources are most divergent, \ie, contribute most to the overall RTD value. Figure.~\ref{fig:sourcertds} illustrates the most divergent sources for different search engines. Here we focus on the top 10 most divergent sources. The comparison between Google and Bing reveals notable differences in the prioritization of sources. Sources such as NPR, Pew Research, CNN, and The Guardian are more likely to appear in Google's results, while Bing shows a strong preference for MSN, likely because MSN and Bing are both owned by the same parent company. \rev{For instance, MSN's RTD score of +0.82 in Bing versus Google indicates it ranks substantially higher in Bing's results.} A similar pattern emerges when comparing Google with DuckDuckGo. In contrast, the comparison between Bing and DuckDuckGo reveals less pronounced divergence. Despite this smaller gap, subtle differences in content surfacing can still be observed. 

This variation underscores the distinct curation strategies employed by search engines, which can influence the diversity of information available to users depending on their chosen platform.

% \ap{this is aggregate; do we need to divide this up; break into leanings and stuf?}

\begin{table}[t]
\centering
\caption{Unique Sources by Search Engine and Query Leaning}
\vspace{-.3cm}
\label{tab:crosstab}
\begin{tabular}{lccc|c}
\toprule
\textbf{Search Engine} & \textbf{Left} & \textbf{Right} & \textbf{Neutral} & \textbf{Total} \\
\midrule
Bing News         & 172 & 159 & 191 & 522 \\
DuckDuckGo News   & 273 & 249 & 308 & 830 \\
Google News       & 222 & 198 & 274 & 694 \\
\midrule
\textbf{Total}     & 667 & 606 & 773 & 2046 \\
\bottomrule
\end{tabular}
\end{table}

Table~\ref{tab:crosstab} presents the number of unique sources by search engine and political leaning\footnote{Note that sources are not disjoint across conditions, so the totals in Table~\ref{tab:crosstab} do not represent unique sources across conditions}.  A $\chi^2$ test of independence found significant differences between the number of unique sources between search engines and political leanings \( \chi^2(4, N = 2046) = 1.39, p = 0.85 \). This suggests that there is no significant association in the number of unique sources between search engine and political leanings.

\section{Bias and Factuality across Ideologies}

While research on algorithmic transparency in social media and Web search has been extensive, the relationship between search engines and political ideology remains underexplored. This section investigates (\textbf{RQ4}) how ideological bias influences the prioritization of sources in top search results, examining whether and how search engines curate results based on political leanings and how this affects the balance between bias and factuality.

\subsection{Source Ideological Bias}

\iffalse
\begin{figure}[t]
\input{figures/density}
\caption{Density Distribution of News Articles by Political Leaning and Factuality Across Search Engines. All three search engines predominantly return news from left-center and highly factual sources.}
\label{fig:density}
\end{figure}
\fi
% \input{figures/density}

\begin{figure}
    \centering
        \begin{tikzpicture}
        \begin{groupplot}[
            group style={group size=1 by 2, vertical sep=0.2cm,},
            height=4.5cm,width=5cm,
            ybar stacked,
            % width=0.4\textwidth,
            % height=6cm,
            ymin=0, ymax=1,
            yticklabels={0\%, 25\%, 50\%, 75\%, 100\%},
            ytick={1.0,0.75, 0.5, 0.25, 0.0},
            xlabel = {},
            ylabel = {},
            xlabel style={font=\small},
            ylabel style={font=\small},
            x tick label style={rotate=45,anchor=east, font=\scriptsize},
            y tick label style={font=\scriptsize},
            enlarge x limits=0.4,
            nodes near coords,
            nodes near coords style={
                font=\tiny,
                color=black,
                /pgf/number format/fixed,
                /pgf/number format/precision=1
            },
            every axis plot/.append style={
                line width=0.2pt,
                draw=black!30 % subtle border
            },
            % legend style={
            %     at={(0.8,-0.65)},
            %     anchor=north,
            %     legend columns=3,
            %     cells={anchor=west},
            %     font=\scriptsize,
            %     draw=none,
            %     row sep=2pt,
            %     column sep=10pt
            % },
            ymajorgrids=true,
            grid style={dashed, gray!30}
        ]
        
        % First plot: By Search Engine
        \nextgroupplot[
            symbolic x coords={Google, Bing, DuckDuckGo},
            xticklabels={},
            bar width=14pt,
            xtick=data,
            y dir = reverse,
            ylabel={MBFC Political Bias},
            xlabel={},    
            legend style={
                at={(1.0,1)},
                anchor=north west,
                legend columns=1,
                cells={anchor=west},
                font=\scriptsize,
                draw=none,
                fill=none,
                row sep=1pt,
                column sep=1pt
            },
        legend entries = {
                Extreme Right,
                Right,
                Right-Center,
                Center,
                Left-Center,
                Left,
                Extreme Left
            }  
        ]
\addplot+[ybar, draw=extremeright, fill=extremeright!40] 
    coordinates {(Google,0.000000) (Bing,0.000000) (DuckDuckGo,0.000000)};

\addplot+[ybar, draw=right, fill=right!40] 
    coordinates {(Google,0.032761) (Bing,0.059014) (DuckDuckGo,0.037093)};  

\addplot+[ybar, draw=rightcenter, fill=rightcenter!40] 
    coordinates {(Google,0.082293) (Bing,0.151172) (DuckDuckGo,0.145344)};

\addplot+[ybar, draw=center, fill=center!40] 
    coordinates {(Google,0.170827) (Bing,0.176233) (DuckDuckGo,0.198335)};

\addplot+[ybar, draw=leftcenter, fill=leftcenter!40] 
    coordinates {(Google,0.584243) (Bing,0.573161) (DuckDuckGo,0.575322)};

\addplot+[ybar, draw=left, fill=left!40] 
    coordinates {(Google,0.129095) (Bing,0.040420) (DuckDuckGo,0.043906)};

\addplot+[ybar, draw=extremeleft, fill=extremeleft!40] 
    coordinates {(Google,0.000780) (Bing,0.000000) (DuckDuckGo,0.000000)};

        % Second plot: By Political Alignment
        \nextgroupplot[
            symbolic x coords={Google, Bing, DuckDuckGo},
            bar width=14pt,
            y dir = reverse,
            xtick=data,
            ylabel={MBFC Factuality},
            xlabel={Search Engine},    
            xlabel style={yshift=6pt},
            legend style={
                at={(1.0,1)},
                anchor=north west,
                legend columns=1,
                cells={anchor=west},
                font=\scriptsize,
                draw=none,
                fill=none,
                row sep=1pt,
                column sep=1pt
            },
    legend entries = {
        Very High,
        High,        
        Mixed,
        Low
    }  
        ]

\addplot+[ybar, draw=veryhigh, fill=veryhigh!40] 
    coordinates {(Google,0.088159) (Bing,0.014521) (DuckDuckGo,0.012262)};

\addplot+[ybar, draw=high, fill=high!40] 
    coordinates {(Google,0.773605) (Bing,0.865440) (DuckDuckGo,0.897820)};

\addplot+[ybar, draw=mixed, fill=mixed!40] 
    coordinates {(Google,0.133020) (Bing,0.105518) (DuckDuckGo,0.089010)};    

\addplot+[ybar, draw=low, fill=low!40] 
    coordinates {(Google,0.005216) (Bing,0.014521) (DuckDuckGo,0.000908)};

        \end{groupplot}

    \end{tikzpicture}
\caption{
Stacked bar charts showing the distribution of MBFC Political Bias (top) and MBFC Source Factuality (bottom) across search engines. Google and DuckDuckGo favor center and left-leaning sources, while Bing returns more right-leaning content. All platforms predominantly surface high-factuality sources.
}
    \label{fig:enter-label}
\end{figure}
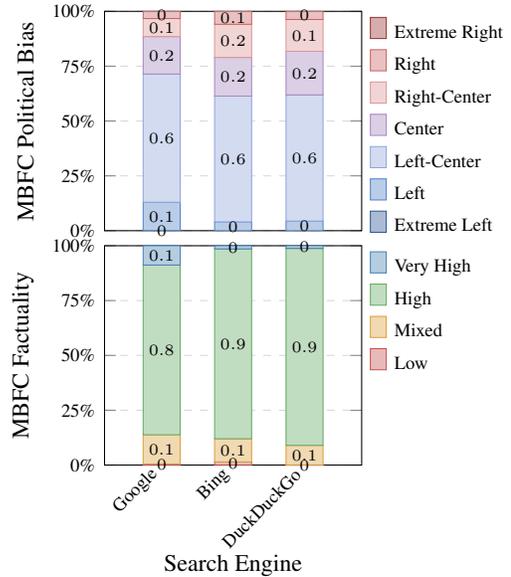

To assess the ideological leaning and factual reliability of retrieved sources, we rely on ratings from Media Bias/Fact Check (MBFC), an independent and widely used news source rating platform~\cite{bozarth2020higher}. MBFC assigns each outlet a political bias label on a 7-point scale ranging from extreme left (-3) to extreme right (+3), and a factuality rating on a 6-point ordinal scale from very low to very high. Sources categorized as conspiracy, satire, or unknown were excluded from quantitative analysis. MBFC ratings have been widely utilized in previous studies to annotate bias and factual accuracy~\cite{starbird2017examining, darwish2017trump} and serve as a ground truth for predictive models~\cite{dinkov2019predicting}.

\rev{Following prior work~\cite{weld2021political}, we converted MBFC’s categorical ratings into numerical scores to enable statistical comparison across search engines and query types.}

\rev{To evaluate differences in source bias and factuality, we applied a combination of statistical methods. We first conducted two-way ANOVA tests with search engine and query leaning as independent variables. Post-hoc Tukey HSD tests were used to assess pairwise differences. To account for the nested structure of our data—multiple results per query—we also fit linear mixed-effects models (LMMs) with random intercepts for each query and fixed effects for search engine, query leaning, and ranking position. This combination of analyses allows us to detect both overall patterns and within-query variation.}

The key questions driving our analysis are: (1) Do search engines amplify or mitigate the ideological leaning of the sources behind the top-ranked content? (2) Are certain search engines more prone to ideological bias than others? (3) How does the interplay between search engine and query leaning affect the political orientation of the retrieved information?

\subsection{Result Bias vs Search Engine and Query Leaning}

\rev{We begin our analysis by examining how the political bias of retrieved news sources varies as a function of search engine and query leaning. Using the MBFC-coded bias scores described above, we fit a two-way ANOVA model with search engine, query leaning, and their interaction as predictors of source bias.}

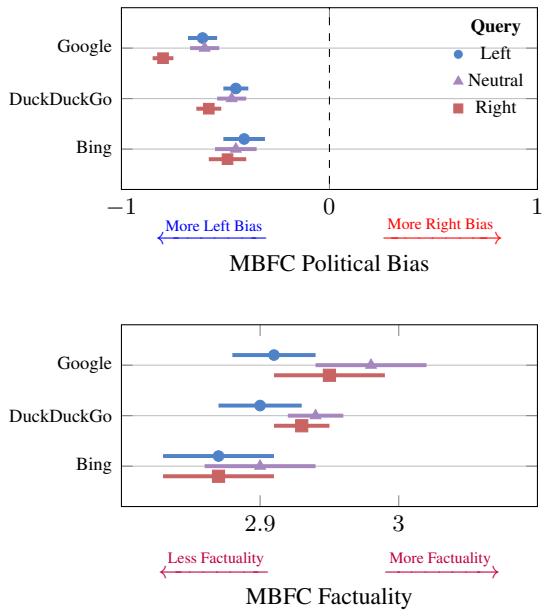
\begin{figure}[t]
    \centering
    %% ---leftFrame ---
\pgfplotstableread{
engine mean ci_lower ci_upper
Bing -0.41 -0.52 -0.31
Duckduckgo -0.45 -0.51 -0.39
Google -0.61 -0.68 -0.54
}\datatableleft

%% ---rightFrame ---
\pgfplotstableread{
engine mean ci_lower ci_upper
Bing -0.49 -0.59 -0.40
Duckduckgo -0.58 -0.64 -0.52
Google -0.80 -0.86 -0.75
}\datatableright

%% ---universalFrame ---
\pgfplotstableread{
engine mean ci_lower ci_upper
Bing -0.45 -0.54 -0.35
Duckduckgo -0.47 -0.53 -0.40
Google -0.60 -0.67 -0.53
}\datatableuniversal

%% ---leftFrame ---
\pgfplotstableread{
engine mean ci_lower ci_upper
Bing 2.87 2.83 2.91
Duckduckgo 2.90 2.88 2.93
Google 2.91 2.87 2.94
}\datatableleftfactuality

%% ---rightFrame ---
\pgfplotstableread{
engine mean ci_lower ci_upper
Bing 2.87 2.82 2.91
Duckduckgo 2.93 2.90 2.95
Google 2.95 2.91 2.99
}\datatablerightfactuality

%% ---universalFrame ---
\pgfplotstableread{
engine mean ci_lower ci_upper
Bing 2.90 2.86 2.94
Duckduckgo 2.94 2.91 2.96
Google 2.98 2.95 3.02
}\datatableuniversalfactuality

\begin{tikzpicture}
\begin{groupplot}[
    group style={
        group size=1 by 2,
        horizontal sep=0.2cm,
        vertical sep=1.8cm,
    },    
    legend style={
        at={(1,1)},
        anchor=north east,
        draw=none,
        fill=none,
        font=\scriptsize,        
    },
    legend cell align={center},
    legend columns=1,
    ymajorgrids=true,
    width=0.4\textwidth,
    height=4cm,
    ymin=-0, ymax=2,
    enlarge y limits=0.4,
    %y dir=reverse,
]

\nextgroupplot[
    ytick={0,1,2},
    xmin=-1, xmax = 1,
    xtick = {-1, 0, 1},
    ylabel={},
    yticklabels={Bing, DuckDuckGo, Google},
    yticklabel style={font=\scriptsize},
    xlabel style = {font=\footnotesize, 
    align=center, yshift=7pt},
    xlabel={\textcolor{blue}{$\xleftarrow{\textrm{More Left Bias}}$}\hspace{1.5cm}\textcolor{red}{$\xrightarrow{\textrm{More Right Bias}}$} \\[0.1pt] MBFC Political Bias},
]

\addplot+[
    only marks,
    mark=*,
    mark options={fill=left, draw=left},
    mark size=2.0pt,
    forget plot,
    error bars/.cd,
        x dir=both,
        x explicit,
        error mark=bar,
        error bar style={line width=1.5pt, draw=left},
] table[
    y expr=\coordindex + 0.2,
    x=mean,
    x error expr=\thisrow{ci_upper} - \thisrow{mean}
] \datatableleft;

\addlegendimage{only marks, mark=o, mark size=0pt, fill=none, draw=none}
\addlegendentry{{\textbf{Query}}}

\addlegendimage{only marks, mark=*, mark size=1.5pt, fill=left, draw=left}
\addlegendentry{Left}

% moderate
\addplot+[
    only marks,
    mark=triangle*,
    mark options={fill=center, draw=center},
    mark size=2.3pt,
    forget plot,
    error bars/.cd,
        x dir=both,
        x explicit,
        error mark=bar,
        error bar style={line width=1.5pt, draw=center},
] table[
    y expr=\coordindex - 0,
    x=mean,
    x error expr=\thisrow{ci_upper} - \thisrow{mean}
] \datatableuniversal;
\addlegendimage{only marks, mark=triangle*, mark size=2.0pt, fill=center, draw=center}
\addlegendentry{Neutral}

\addplot+[
    only marks,
    mark=square*,
    mark options={fill=right, draw=right},
    mark size=2.0pt,
    forget plot,
    error bars/.cd,
        x dir=both,
        x explicit,
        error mark=bar,
        error bar style={line width=1.5pt, draw=right},
] table[
    y expr=\coordindex - 0.2,
    x=mean,
    x error expr=\thisrow{ci_upper} - \thisrow{mean}
] \datatableright;
\addlegendimage{only marks, mark=square*, mark size=1.7pt, fill=right, draw=right}
\addlegendentry{Right}

\draw[dashed] (axis cs:0,-1) -- (axis cs:0,3);

\nextgroupplot[
    xmin=2.8, xmax=3.1,
    xtick={2.9, 3.0},
    ylabel={},
    ylabel style={font=\footnotesize, yshift=15pt},
    ytick={0,1,2},
    yticklabels={Bing, DuckDuckGo, Google},
    yticklabel style={font=\scriptsize},
    xlabel style = {font=\footnotesize, 
    align=center},
    xlabel={\textcolor{purple}{$\xleftarrow{\textrm{Less Factuality}}$}\hspace{1.5cm}\textcolor{purple}{$\xrightarrow{\textrm{More Factuality}}$} \\[0.1pt] MBFC Factuality},
]
\addplot+[
    only marks,
    mark=*,
    mark options={fill=left, draw=left},
    mark size=2.0pt,
    forget plot,
    error bars/.cd,
        x dir=both,
        x explicit,
        error mark=bar,
        error bar style={line width=1.5pt, draw=left},
] table[
    y expr=\coordindex + 0.2,
    x=mean,
    x error expr=\thisrow{ci_upper} - \thisrow{mean}
] \datatableleftfactuality;

% moderate
\addplot+[
    only marks,
    mark=triangle*,
    mark options={fill=center, draw=center},
    mark size=2.3pt,
    forget plot,
    error bars/.cd,
        x dir=both,
        x explicit,
        error mark=bar,
        error bar style={line width=1.5pt, draw=center},
] table[
    y expr=\coordindex - 0,
    x=mean,
    x error expr=\thisrow{ci_upper} - \thisrow{mean}
] \datatableuniversalfactuality;

\addplot+[
    only marks,
    mark=square*,
    mark options={fill=right, draw=right},
    mark size=2.3pt,
    forget plot,
    error bars/.cd,
        x dir=both,
        x explicit,
        error mark=bar,
        error bar style={line width=1.5pt, draw=right},
] table[
    y expr=\coordindex - 0.2,
    x=mean,
    x error expr=\thisrow{ci_upper} - \thisrow{mean}
] \datatablerightfactuality;

\end{groupplot}
\end{tikzpicture}
    \caption{Mean source bias (top) and factuality (bottom) by search engine and query leaning. Bars show 95\% confidence intervals. While DuckDuckGo and Bing return more right-leaning sources than Google, Tukey tests reveal that right-leaning queries unexpectedly retrieve more left-leaning content. Google also consistently returns higher factuality sources than Bing, with neutral queries producing the most reliable results across platforms.}
    \label{fig:rq4_tukey}
\end{figure}

\rev{The ANOVA revealed significant main effects of both search engine ($F(2, 5143) = 30.69$, $p < .001$) and political query leaning ($F(2, 5143) = 15.01$, $p < .001$) on the bias scores of returned sources. The interaction between search engine and leaning was not significant, suggesting that the effect of political leaning on source bias is consistent across platforms.}

\rev{Post-hoc Tukey tests, illustrated in Fig.~\ref{fig:rq4_tukey}, showed that Google returns significantly less biased sources than both Bing ($p < .001$) and DuckDuckGo ($p < .001$), while Bing and DuckDuckGo do not significantly differ from one another. In terms of query leaning, right-leaning queries are associated with more ideologically extreme sources than both left-leaning ($p < .001$) and neutral queries ($p < .001$), while no significant difference was found between left-leaning and neutral queries ($p = .99$).}

\rev{Overall, these results demonstrate that both platform and query ideology systematically shape the political slant of search results, with Google tending toward centrist sources and right-leaning queries drawing more partisan content across engines.}

\rev{\paragraph{Mixed Effects Model} The preceding ANOVA and Tukey tests assume that all observations are independent, but multiple results are returned for each query, introducing within-query dependencies. To account for this structure, we fit a linear mixed-effects model (LMM) with fixed effects for search engine, query leaning, and result ranking, and a random intercept for each query.}

\rev{We compared a main-effects model to one that included interaction terms between search engine and query leaning. A likelihood ratio test found no improvement in fit from the interaction terms ($\chi^2$(4) = -10.78, $p = 1.00$). We therefore report results from the more parsimonious main-effects model.}

\rev{This model confirmed our earlier findings: Bing and DuckDuckGo return significantly more right-leaning sources than Google, and right-leaning queries are associated with more partisan content compared to neutral queries. Left-leaning queries do not differ significantly from neutral ones, and ranking position had no measurable effect on bias. A random intercept for each query accounted for within-query dependencies. Full model coefficients are presented in Fig.~\ref{fig:lmm}.}

\rev{The ideological leaning of the user’s query consistently influenced the bias of retrieved sources. Across all models, \textbf{right-leaning queries were associated with a clear shift toward more left-leaning sources}---more so than either neutral or left-leaning queries. This counterintuitive pattern may reflect algorithmic efforts by search engines to balance or counteract the ideological slant of the query itself. Alternatively, it may be shaped by the dynamics of \textit{negative partisanship}, wherein media—particularly left-leaning outlets—devote more coverage to political opponents than to ideological allies~\cite{abramowitz2018negative}. For example, Republican-leaning sources may publish more critical content about Biden than Democratic sources publish about Trump.}

\subsection{Result Factuality vs Search Engine and Query Leaning}

\rev{To assess how search engines and query leaning influence the reliability of retrieved information, we assigned ordinal scores to MBFC factuality ratings, ranging from 1 (low) to 5 (very high). A two-way ANOVA revealed significant main effects of both search engine ($F(2, 5143) = 9.05$, $p < .001$) and query leaning ($F(2, 5143) = 5.89$, $p = .003$), with no significant interaction. }

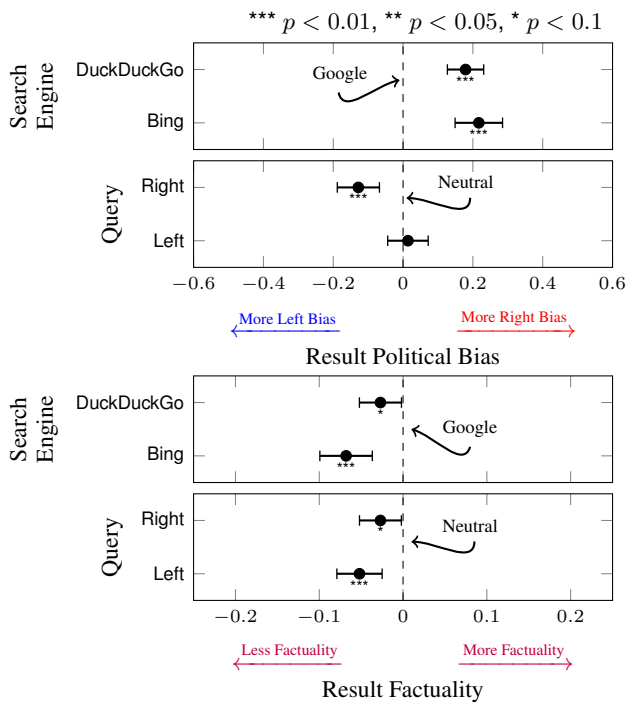
\begin{figure}[t]
    \begin{tikzpicture}
\begin{groupplot}[
    group style={
        group name=coefplot,
        group size=1 by 2, % 1 column, 3 rows
        vertical sep=0.15cm, % Vertical separation between plots
    },
    width=.85\linewidth,
    xlabel style = {align=center, font=\footnotesize},
    xlabel={\textcolor{blue}{$\xleftarrow{\textrm{More Left Bias}}$}\hspace{1.5cm}\textcolor{red}{$\xrightarrow{\textrm{More Right Bias}}$} \\ Result Political Bias},
    ylabel={},
    ylabel style = {font=\sffamily\footnotesize},
    yticklabel style = {font=\sffamily\scriptsize},
    xticklabel style = {font=\sffamily\scriptsize},
    xmax = 0.6,
    xmin = -0.6,
    legend style={
        at={(0.35,-0.25)},
        anchor=north,
        legend columns=5,
        cells={anchor=west},
    },
    scatter/classes={
    conflict={mark=*,red}, 
    economic={mark=*,blue}, 
    episodic={mark=*,teal}, 
    game={mark=*,orange},
    human={mark=*,black}
    },
    scatter,
    clip=false
]

\nextgroupplot[
    %title=First Plot Title,
        ytick={1,2},
        yticklabels={
            Bing,
            DuckDuckGo
        },
        ylabel={},
        ymin = 0.5,
        ymax = 2.5,
        height=3cm,
        only marks,
        ylabel={Search\\Engine},
        ylabel style = {align=center, font=\footnotesize},
        xticklabels={},
        xlabel={},      % Suppress x-axis label
]
    % Vertical reference line
    \addplot[dashed, samples=50, smooth, no marks, domain=0:10, black] coordinates {(0,0.5)(0,2.5)};
    %\node[anchor=south east, font=\footnotesize\sffamily] at (axis cs:0.6,2.5) {*** $p<0.01$, ** $p<0.05$, * $p<0.1$};

\node[anchor=south east, font=\footnotesize\sffamily] at (axis cs:0.6,2.5) {*** $p<0.01$, ** $p<0.05$, * $p<0.1$};

    \node[black] (baseline) at (axis cs:-0.18,1.9) {\scriptsize{Google}};
    \draw[->, bend right=30, thick] (baseline) to[out=-90, in=180] (axis cs:-0.01,1.8);

    % Error bars with y-shift
    \addplot [
        scatter src=explicit symbolic,
        visualization depends on={value \thisrow{lab} \as \Label},
        nodes near coords*={\tiny{\sffamily{\Label}}},
        every node near coord/.append style={
            anchor=north, % Align the text to the west of the node
            yshift =1pt
        },
        error bars/.cd, 
        x dir = both, 
        x explicit, 
        error bar style={thick, solid, black},
        error mark options={line width=0.5pt, mark size=2pt, rotate=90}
    ]    
    table[meta=class, x=x, y=y, x error=ex]{
        y x         ex          class   lab     
        1 0.217   0.068     d ***
        2 0.179   0.052     d ***
};
\nextgroupplot[
        ytick={3,4},
        yticklabels={
            Left, Right
        },
        ylabel={Query},
        ylabel style = {align=center, font=\footnotesize},
        ymin = 2.5,
        ymax = 4.5,  
        height=3cm,
        only marks,
        title={},
]
    % Vertical reference line
    \addplot[dashed, samples=50, smooth, no marks, domain=0:10, black] coordinates {(0,2.5)(0,4.5)};

    \node[black] (baseline) at (axis cs:0.18,4.1) {\scriptsize{Neutral}};
    \draw[->, bend right=30, thick] (baseline) to[out=90, in=180] (axis cs:0.01,3.8);

    % Error bars with y-shift
    \addplot [        
        scatter src=explicit symbolic,
        visualization depends on={value \thisrow{lab} \as \Label},
        nodes near coords*={\tiny{\sffamily{\Label}}},
        every node near coord/.append style={
            anchor=north, % Align the text to the west of the node
            yshift =1pt
        },
        error bars/.cd, 
        x dir = both, 
        x explicit, 
        error bar style={thick, solid, black},
        error mark options={line width=0.5pt, mark size=2pt, rotate=90}
    ]
    table[meta=class, x=x, y=y, x error=ex]{
        y x         ex          class   lab        
        3 0.014    0.058     d {}
        4 -0.128    0.06038     d ***        
        };
    \end{groupplot}
\end{tikzpicture}

\begin{tikzpicture}
\begin{groupplot}[
    group style={
        group name=coefplot,
        group size=1 by 2, % 1 column, 3 rows
        vertical sep=0.15cm, % Vertical separation between plots
    },
    width=.85\linewidth,
    xlabel style = {align=center, font=\footnotesize},
    xlabel={\textcolor{purple}{$\xleftarrow{\textrm{Less Factuality}}$}\hspace{1.5cm}\textcolor{purple}{$\xrightarrow{\textrm{More Factuality}}$} \\ Result Factuality },
    ylabel={},
    ylabel style = {font=\sffamily\footnotesize},
    yticklabel style = {font=\sffamily\scriptsize},
    xticklabel style = {font=\sffamily\scriptsize},
    xmax = 0.25,
    xmin = -0.25,
    legend style={
        at={(0.35,-0.25)},
        anchor=north,
        legend columns=5,
        cells={anchor=west},
    },
    scatter/classes={
    conflict={mark=*,red}, 
    economic={mark=*,blue}, 
    episodic={mark=*,teal}, 
    game={mark=*,orange},
    human={mark=*,black}
    },
    scatter,
    clip=false
]

\nextgroupplot[
    %title=First Plot Title,
        ytick={1,2},
        yticklabels={
            Bing,
            DuckDuckGo
        },
        ylabel={Search\\Engine},
        ylabel style = {align=center, font=\footnotesize},
        ymin = 0.5,
        ymax = 2.5,
        height=3cm,
        only marks,
        xticklabels={},
        xlabel={},      % Suppress x-axis label
]
    % Vertical reference line
    \addplot[dashed, samples=50, smooth, no marks, domain=0:10, black] coordinates {(0,0.5)(0,2.5)};

    \node[black] (baseline) at (axis cs:0.08,1.5) {\scriptsize{Google}};
    \draw[->, bend right=30, thick] (baseline) to[out=90, in=180] (axis cs:0.01,1.5);

    % Error bars with y-shift
    \addplot [
        scatter src=explicit symbolic,
        visualization depends on={value \thisrow{lab} \as \Label},
        nodes near coords*={\tiny{\sffamily{\Label}}},
        every node near coord/.append style={
            anchor=north, % Align the text to the west of the node
            yshift =1pt
        },
        error bars/.cd, 
        x dir = both, 
        x explicit, 
        error bar style={thick, solid, black},
        error mark options={line width=0.5pt, mark size=2pt, rotate=90}
    ]
    table[meta=class, x=x, y=y, x error=ex]{
        y x         ex          class   lab        
        1 -0.068  0.0313      d {***}
        2 -0.027  0.0251      d {*}
};
\nextgroupplot[
        ytick={3,4},
        yticklabels={
            Left, Right
        },
        ymin = 2.5,
        ymax = 4.5,  
        ylabel={Query},
        ylabel style = {align=center, font=\footnotesize},
        height=3cm,
        only marks,
        title={},
]
    % Vertical reference line
    \addplot[dashed, samples=50, smooth, no marks, domain=0:10, black] coordinates {(0,2.5)(0,4.5)};

    \node[black] (baseline) at (axis cs:0.08,3.9) {\scriptsize{Neutral}};
    \draw[->, bend right=30, thick] (baseline) to[out=90, in=180] (axis cs:0.01,3.6);

    % Error bars with y-shift
    \addplot [        
        scatter src=explicit symbolic,
        visualization depends on={value \thisrow{lab} \as \Label},
        nodes near coords*={\tiny{\sffamily{\Label}}},
        every node near coord/.append style={
            anchor=north, % Align the text to the west of the node
            yshift =1pt
        },
        error bars/.cd, 
        x dir = both, 
        x explicit, 
        error bar style={thick, solid, black},
        error mark options={line width=0.5pt, mark size=2pt, rotate=90}
    ]
    table[meta=class, x=x, y=y, x error=ex]{
        y x         ex          class   lab            
        3 -0.052  0.027      d ***
        4 -0.027  0.025      d * 
        };
    \end{groupplot}
\end{tikzpicture}
\caption{Fixed effects estimates from linear mixed-effects models predicting result bias (top) and result factuality (bottom). Error bars show 95\% confidence intervals. Results confirm that Google returns more left-leaning and higher-factuality sources than other platforms. Right-leaning queries are associated with more partisan sources, though not lower factuality. Ranking position was not a significant predictor in either model.}
    \label{fig:lmm}
\end{figure}

\rev{Tukey post-hoc comparisons indicate that Google returns significantly higher-factuality sources than both Bing ($p < .001$) and DuckDuckGo ($p = .018$). Bing and DuckDuckGo also differ slightly ($p = .018$), though to a lesser degree. Among query types, neutral queries yielded more factual sources than left-leaning queries ($p = .003$), while differences between right-leaning queries and the other groups were not statistically significant.}

\rev{\paragraph{Mixed Effects Model} To again account for the nested structure of the data (multiple results per query), we also fit a linear mixed-effects model with fixed effects for search engine, query leaning, and ranking, and a random intercept for each query key. Results from this model mirror the ANOVA findings: Google surfaces the most factual content, followed by DuckDuckGo and Bing. Left-leaning queries produced marginally less factual results than neutral queries, while right-leaning queries showed a borderline-significant decrease. Ranking position had no measurable effect on source quality. Full model estimates are shown in Fig.~\ref{fig:lmm} (bottom).}

\rev{Taken together, these findings indicate that both platform and query framing subtly shape the factuality of retrieved sources. Neutral queries consistently yield the most reliable information across platforms. While differences across search engines are small, they are directionally consistent: Google provides a slight edge in factual accuracy, especially compared to DuckDuckGo. Importantly, politically charged queries (whether left- or right-leaning) tend to surface less factual sources than neutral queries, underscoring how user framing influences not just the ideological content but also the quality of results.}

\section{Discussion}
Our study examines how major news search engines---Google News, Bing News, and DuckDuckGo News---mediate access to politically inclined queries. We analyze semantic polarity, framing effects, source selection, and ideological bias and factuality to understand their influence on the online news ecosystem and public discourse.

Neutral queries act as bridges, offering semantically diverse results that promote balanced perspectives crucial for democracy. In contrast, politically charged queries exhibit greater polarization, often reinforcing ideological silos. Notably, Google News mitigates this effect better than Bing News and DuckDuckGo News.  

Framing analysis reveals that Bing and DuckDuckGo prioritize conflict and strategy narratives, especially for right-leaning queries, potentially deepening partisan divides. These tendencies influence how users interpret and emotionally engage with news stories~\cite{entman1993framing}, heightening perceptions of political contention while reducing focus on substantive issues.  

Our source-level analysis shows that Google prioritizes outlets like NPR and CNN, while Bing and DuckDuckGo exhibit more overlap with each other and include a broader range of partisan and lesser-known domains. \rev{This divergence may stem from Google's greater investment in algorithmic trustworthiness metrics or its different editorial standards.} While it is unsurprising that different platforms surface different sources, the consequences are nontrivial: \rev{search engine design choices systematically shape what counts as visible, credible news in response to a user’s query.}%Exposure to less factual information can misinform the public, hinder democratic processes, and exacerbate polarization~\cite{lazer2018science}.  

\rev{One especially noteworthy finding is that right-leaning queries tend to retrieve more left-leaning sources. While this could reflect algorithmic efforts to balance perceived user bias, it also aligns with theories of negative partisanship, where partisan media disproportionately focus on critiquing the opposition. This suggests that even oppositional coverage can amplify attention to certain political actors, shaping perceptions indirectly.}

In terms of factuality, DuckDuckGo returned consistently high-factuality sources across queries, while Google’s broader inclusion criteria produced more variability. \rev{This highlights a subtle tradeoff: platforms that emphasize diversity and breadth may also expose users to more mixed-quality information. Designing for both balance and credibility remains a persistent challenge.}

\rev{Taken together, our findings reveal how search engines act not just as gatekeepers but as agenda-setters, shaping what users see, how it is framed, and whose voices are amplified. These effects are not accidental—they arise from the structure of search algorithms, source selection heuristics, and response to ideological cues embedded in queries.}

\rev{Our results suggest three concrete implications. First, encouraging neutral or descriptive search behavior could reduce polarization in news exposure. Second, platform-level decisions about source and frame prioritization have downstream effects on users’ political understanding. Third, designers and policymakers should treat algorithmic transparency and ideological fairness not as abstract goals but as measurable, testable system behaviors that can be audited and improved.}

\section{Conclusions}

In conclusion, we demonstrate that search engines not only mediate access to information but also actively shape the nature of political discourse by influencing semantic exposure, framing, source diversity, and factuality. The variations among Google News, Bing News, and DuckDuckGo News reflect differing algorithmic approaches with significant implications for ideological polarization and informed citizenship.

As digital platforms continue to be primary gateways to information, it is imperative for search engines to prioritize transparency, promote content diversity, and uphold high standards of factual accuracy. Stakeholders—including platform developers, policymakers, and educators—must collaborate to ensure that the design and operation of these algorithms serve the public interest by facilitating access to reliable and diverse information. Only through such concerted efforts can we mitigate the risks of echo chambers and support a vibrant, informed, and democratic society.

\paragraph{Limitations} \rev{While our study offers insights into how news search engines mediate politically inclined queries, several limitations merit attention. First, data were collected in late September 2024, during the lead-up to the U.S. presidential election. This timing is appropriate for capturing politically salient dynamics but may limit generalizability to less active political periods. In addition, we did not constrain result publication dates, so some headlines may predate the snapshot. Future research should adopt longitudinal designs to examine how platform responses evolve over time, especially around major events.}

\rev{Second, our analysis focuses on English-language queries in the U.S., limiting its applicability to other regions and linguistic contexts. Expanding to multilingual and cross-national comparisons would offer a more global view of how search engines curate political information.}

\rev{Third, our framing measures rely only on the headline that appears on the search‑results page. This choice reflects the practical reality that these cues are all most users see before deciding whether to click. Nonetheless, restricting the analysis to surface text may miss frames or partisan nuance embedded in the full article. Headlines often compress or soften the ideological stance found deeper in the story, and some outlets craft headlines for engagement rather than comprehensive framing. Future work should incorporate fuller article context, \eg, first‑paragraph or entire‑text embeddings, where licensing permits, to assess when headline framing diverges from in‑depth content and how that affects user interpretation.}

% Third, we selected contentious topics for our queries, building on previous research that highlights their relevance in examining ideological bias and framing effects. While this focus enhances the study's relevance to discussions on polarization, it may not reflect the full spectrum of topics that users engage with. Future work should consider a broader range of subjects, including less contentious or non-political topics, to determine whether the observed patterns hold across different content areas.

% Moreover, our study did not account for user-specific factors such as personalization based on individual search history, location nuances beyond the set parameters, or device type. Search engines often tailor results to individual users, which can significantly influence the information landscape each user encounters. Incorporating these personalization factors in future research would provide a more nuanced understanding of how individual experiences vary and how personalization may contribute to information diversity or reinforcement of existing beliefs.

% Additionally, while we utilized Media Bias/Fact Check (MBFC) as our tool for assessing media bias and factuality—chosen for its extensive coverage and validation in prior research—reliance on a single assessment tool may introduce limitations. Future studies should consider integrating multiple bias and factuality assessment frameworks to cross-validate findings and account for potential discrepancies between different evaluation methodologies.

Finally, although our research sheds light on the content presented by search engines, it does not measure the priming effects of framing, sources, or ideological bias on user opinions and public discourse. Understanding the influence of these factors on individual attitudes and societal polarization requires experimental approaches. Future work could involve controlled experiments or surveys to assess how exposure to specific frames or biased sources affects opinion shifts, providing deeper insights into the causal relationships between search engine content and public perceptions.

By addressing these limitations, subsequent research can build upon our findings to enhance the understanding of search engines' roles in shaping information access and the broader implications for democratic engagement and public discourse.

% \begin{figure}[t]
%     \input{data/fact}
%     \caption{(A) OLR coefficients for source factuality and their standard errors according to MBFC are plotted on the x-axis. Google is the coded baseline for the search engines; Neutral is the coded baseline for the query leanings. (B) Posthoc pairwise comparisons with Tukey's method. }
%     \label{fig:factuality_olr}
% \end{figure}

\bibliography{aaai25}

\begin{enumerate}

\item For most authors...
\begin{enumerate}
    \item  Would answering this research question advance science without violating social contracts, such as violating privacy norms, perpetuating unfair profiling, exacerbating the socio-economic divide, or implying disrespect to societies or cultures?
    \answerYes{Yes}
  \item Do your main claims in the abstract and introduction accurately reflect the paper's contributions and scope?
    \answerYes{Yes}
   \item Do you clarify how the proposed methodological approach is appropriate for the claims made? 
    \answerYes{Yes}
   \item Do you clarify what are possible artifacts in the data used, given population-specific distributions?
    \answerYes{Yes}
  \item Did you describe the limitations of your work?
    \answerYes{Yes}
  \item Did you discuss any potential negative societal impacts of your work?
    \answerNA{NA}
      \item Did you discuss any potential misuse of your work?
    \answerNA{NA}
    \item Did you describe steps taken to prevent or mitigate potential negative outcomes of the research, such as data and model documentation, data anonymization, responsible release, access control, and the reproducibility of findings?
     \answerNA{NA}
  \item Have you read the ethics review guidelines and ensured that your paper conforms to them?
     \answerYes{Yes}
\end{enumerate}

\item Additionally, if your study involves hypotheses testing...
\begin{enumerate}
  \item Did you clearly state the assumptions underlying all theoretical results?
   \answerNA{NA}
  \item Have you provided justifications for all theoretical results?
    \answerNA{NA}
  \item Did you discuss competing hypotheses or theories that might challenge or complement your theoretical results?
    \answerNA{NA}
  \item Have you considered alternative mechanisms or explanations that might account for the same outcomes observed in your study?
    \answerNA{NA}
  \item Did you address potential biases or limitations in your theoretical framework?
    \answerNA{NA}
  \item Have you related your theoretical results to the existing literature in social science?
    \answerNA{NA}
  \item Did you discuss the implications of your theoretical results for policy, practice, or further research in the social science domain?
    \answerNA{NA}
\end{enumerate}

\item Additionally, if you are including theoretical proofs...
\begin{enumerate}
  \item Did you state the full set of assumptions of all theoretical results?
   \answerNA{NA}
	\item Did you include complete proofs of all theoretical results?
    \answerNA{NA}
\end{enumerate}

\item Additionally, if you ran machine learning experiments...
\begin{enumerate}
  \item Did you include the code, data, and instructions needed to reproduce the main experimental results (either in the supplemental material or as a URL)?
    \answerNA{NA}
  \item Did you specify all the training details (e.g., data splits, hyperparameters, how they were chosen)?
    \answerNA{NA}
     \item Did you report error bars (e.g., with respect to the random seed after running experiments multiple times)?
    \answerNA{NA}
	\item Did you include the total amount of compute and the type of resources used (e.g., type of GPUs, internal cluster, or cloud provider)?
    \answerNA{NA}
     \item Do you justify how the proposed evaluation is sufficient and appropriate to the claims made? 
    \answerNA{NA}
     \item Do you discuss what is ``the cost`` of misclassification and fault (in)tolerance?
    \answerNA{NA}
  
\end{enumerate}

\item Additionally, if you are using existing assets (e.g., code, data, models) or curating/releasing new assets, \textbf{without compromising anonymity}...
\begin{enumerate}
  \item If your work uses existing assets, did you cite the creators?
   \answerYes{Yes}
  \item Did you mention the license of the assets?
   \answerYes{Yes}
  \item Did you include any new assets in the supplemental material or as a URL?
    \answerNA{NA}
  \item Did you discuss whether and how consent was obtained from people whose data you're using/curating?
    \answerNA{NA}
  \item Did you discuss whether the data you are using/curating contains personally identifiable information or offensive content?
    \answerNA{NA}
\item If you are curating or releasing new datasets, did you discuss how you intend to make your datasets FAIR (see \citet{fair})?
\answerNA{NA}
\item If you are curating or releasing new datasets, did you create a Datasheet for the Dataset (see \citet{gebru2021datasheets})? 
\answerNA{NA}
\end{enumerate}

\item Additionally, if you used crowdsourcing or conducted research with human subjects, \textbf{without compromising anonymity}...
\begin{enumerate}
  \item Did you include the full text of instructions given to participants and screenshots?
  \answerNA{NA}
  \item Did you describe any potential participant risks, with mentions of Institutional Review Board (IRB) approvals?
    \answerNA{NA}
  \item Did you include the estimated hourly wage paid to participants and the total amount spent on participant compensation?
   \answerNA{NA}
   \item Did you discuss how data is stored, shared, and deidentified?
   \answerNA{NA}
\end{enumerate}

\end{enumerate}

% \bibliography{aaai25}

% \end{document}

% \clearpage 

\appendix

\renewcommand{\thefigure}{A\arabic{figure}}
\setcounter{figure}{0}
\renewcommand{\thesection}{A\arabic{section}}
\setcounter{section}{0}

\section{Appendix}
\label{appendix}

\subsection{Rank Turbulence Divergence (RTD)}

\label{sec:rtd_eq}

\paragraph{Intuition.}
RTD measures how much two ranked lists disagree, with an emphasis on the highest‑ranked items that users are most likely to see.
An RTD of $0$ means the lists are identical, whereas values approaching $1$ signal that the items that rank highly in one list are largely absent (or much lower) in the other.
The \emph{signed} RTD we report assigns a positive sign when the top ranks in list~$R_1$ shift downward in list~$R_2$ (and a negative sign for the reverse), allowing us to say which system elevates or demotes content.

\paragraph{Formal definition.}
Let $R_1$ and $R_2$ be two ranked lists of tokens (here, news‑source domains) and let
$r_{\xi,1}$ and $r_{\xi,2}$ denote the rank of token~$\xi$ in those lists.
Following \citet{dodds2023allotaxonometry}, we first compute an element‑level divergence
\begin{equation}
\left|\frac{1}{[r_{\xi, 1}]^\alpha}- \frac{1}{[r_{\xi,2}]^\alpha}\right|^{\frac{1}{\alpha+1}}
\label{inverse_control}
\end{equation}
% \vspace{-0.5}
\noindent where the control parameter $\alpha\!\in\!(0,1)$ tunes how strongly early
ranks are weighted (we use $\alpha=\frac{1}{3}$, a common default that balances head and tail items).

The aggregate RTD is then
\begin{equation}
\begin{aligned}
&RTD^R_\alpha(R1 \parallel R2)  \\
&= \frac{1}{N_{1,2;\alpha}} \frac{\alpha+1}{\alpha} \sum\limits_{\xi \in R_{1,2; \alpha}}
\left|\frac{1}{[r_{\xi, 1}]^\alpha} - \frac{1}{[r_{\xi, 2}]^\alpha}\right|^{\frac{1}{\alpha+1}}
\end{aligned}
\label{rtd}
\end{equation}
\noindent where $N_{1,2;\alpha}$ is a normalization factor ensuring
$0\!\le\!RTD_\alpha\!\le\!1$ and
$R_{1,2;\alpha}$ is the union of tokens present in either list.

\paragraph{Concrete example.}
\rev{Consider the domain \texttt{msn.com}, ranked $\,r_{\text{\tiny msn},1}=5$ in Bing News and $\,r_{\text{\tiny msn},2}=20$ in Google News. With $\alpha=\frac{1}{3}$,
$$ \delta_{\tiny 1/3}(\texttt{msn})
  = \bigl|\,5^{-1/3} - 20^{-1/3}\bigr|^{3/4}
  \;\approx\; 0.28. $$
Summing analogous terms over all domains and applying the normalisation yields
a \emph{signed} RTD of $+0.82$ for Bing~vs.~Google,
indicating that Bing surfaces a markedly different—and, on average, higher‑ranked—set of domains than Google for the same query.
For reference, RTD values in our study range from $0.05$ (near‑identical rankings) to $0.86$ (highly turbulent).}

\paragraph{Interpretation scale.}\rev{
\begin{itemize}[leftmargin=1.7em,itemsep=2pt]
  \item \textbf{0\,–\,0.2}: Lists are almost identical.  
  \item \textbf{0.2\,–\,0.5}: Moderate re‑ordering; some head items shift ranks.  
  \item \textbf{0.5\,–\,1.0}: Substantial turbulence; head items in one list are low‑rank or absent in the other.
\end{itemize}
}

\subsection{Prompting Template for Framing Classification}
\label{sec:prompt}

We use the following prompt to classify the given headline into one of the six frames: Conflict, Game/Strategy, Thematic/Issue, Human Interest, Episodic, and Economic. Please find the descriptions to each of the frames in the prompt below.

% Instruction: \\
\texttt{
instruction: You are an assistant trained to classify news headlines into one of the following generic frames based on their dominant focus. Below are the definitions of each frame: \\
1. Conflict Frame \\
   - Description: Presents events as a conflict between competing actors, issues, or interpretations.\\
2. Game/Strategy Frame\\
   - Description: Focuses on the efforts of actors to gain support, influence, or achieve specific goals.\\
3. Thematic/Issue Frame\\
   - Description: Centers on the substantive content of public concerns and issues. \\
4. Human Interest Frame\\
   - Description: Narrates events from the perspective of individuals affected by the issues or events.\\
5. Episodic Frame\\
   - Description: Presents specific events or episodes without extensive context or connection to broader themes.\\
6. Economic Consequences Frame\\
   - Description: Assesses the expected implications of events and policies on the economy.\\
Task:\\
Given a news headline, classify it into one of the above frames by selecting the most appropriate single frame that best represents the headline's primary focus**. Respond only with the name of the frame (e.g., "Conflict Frame").
Additional Notes:\\
- Select Only One Frame: Assign only the dominant frame that best fits the headline, even if multiple frames seem relevant.\\
- Consistency: Use the exact frame names provided in the definitions for clarity and consistency.\\
- Clarity: Ensure that the classification is based solely on the headline's content without requiring external context.\\
Example:\\
- Headline: "Local Hero Rescues Family from Burning Building"\\
- Frame: Human Interest Frame\\
Now, classify the following headline:
\noindent Headline: [headline]\\
}

\clearpage

\end{document}